\documentclass[manuscript, nonacm]{acmart}

\AtBeginDocument{%
  \providecommand\BibTeX{{%
    \normalfont B\kern-0.5em{\scshape i\kern-0.25em b}\kern-0.8em\TeX}}}

\usepackage{epigraph}
\usepackage{csquotes}
\usepackage{amsmath}
\usepackage{amsthm}
\usepackage{orcidlink}
\usetikzlibrary{arrows}
\usetikzlibrary{shapes.geometric}
\usepackage{caption}
\usepackage{subcaption}

\theoremstyle{remark}

\begin{document}

\title{Mutatis Mutandis: Revisiting the Comparator in Discrimination Testing}

\author{Jos\'e M. \'Alvarez \orcidlink{0000-0001-9412-9013}}
\email{jm.alvarez.colmenares@gmail.com}
% \orcid{0000-0001-9412-9013}
\affiliation{
  \institution{Santander AI Lab}
  \city{Madrid}
  \country{Spain}
}

\author{Salvatore Ruggieri \orcidlink{0000-0002-1917-6087}}
\email{salvatore.ruggieri@unipi.it}
% \orcid{0000-0002-1917-6087}
\affiliation{
  \institution{University of Pisa}
  \city{Pisa}
  \country{Italy}
}

\renewcommand{\shortauthors}{\'{A}lvarez and Ruggieri}
\renewcommand{\shorttitle}{Mutatis Mutandis}

\begin{abstract}
    Testing for individual discrimination involves deriving a profile, the comparator, similar to the one making the discrimination claim, the complainant, based on a protected attribute, such as race or gender, and comparing their decision outcomes. The complainant-comparator pair is central to discrimination testing. Most discrimination testing tools rely on this pair to establish evidence for discrimination. In this work, we revisit the role of the comparator in discrimination testing. We first argue for the inherent causal modeling nature of deriving the comparator. We then introduce a two-kind classification for the comparator: the \textit{ceteris paribus}, or ``with all else equal,'' (CP) comparator and the \textit{mutatis mutandis}, or ``with the appropriate adjustments being made,'' (MM) comparator. The CP comparator is the standard comparator, representing an idealized comparison for establishing discrimination as it aims for a complainant-comparator pair that only differs in membership in the protected attribute. As an alternative to the CP comparator, we define the MM comparator, which requires a comparator that represents the ``what would have been'' of the complainant without the effects of the protected attribute on the non-protected attributes. Under the MM comparator, the complainant-comparator pair can be dissimilar in terms of the non-protected attributes, departing from the idealized comparison imposed by the CP comparator. Notably, the MM comparator denotes a more complex object and its implementation offers an impactful venue for machine learning methods. We illustrate these two comparators and their impact on discrimination testing using a real-world example.
\end{abstract}

% %% The code below is generated by the tool at http://dl.acm.org/ccs.cfm.
% \begin{CCSXML}
% <ccs2012>
% <concept>
% <concept_id>10010147.10010257</concept_id>
% <concept_desc>Computing methodologies~Machine learning</concept_desc>
% <concept_significance>500</concept_significance>
% </concept>
% <concept>
% <concept_id>10002950.10003648.10003649</concept_id>
% <concept_desc>Mathematics of computing~Probabilistic representations</concept_desc>
% <concept_significance>500</concept_significance>
% </concept>
% </ccs2012>
% \end{CCSXML}
% \begin{CCSXML}
% <ccs2012>
%    <concept>
%        <concept_id>10010147.10010178.10010187</concept_id>
%        <concept_desc>Computing methodologies~Knowledge representation and reasoning</concept_desc>
%        <concept_significance>500</concept_significance>
%        </concept>
%  </ccs2012>
% \end{CCSXML}

% \ccsdesc[500]{Computing methodologies~Knowledge representation and reasoning}
% \ccsdesc[500]{Computing methodologies~Machine learning}
% \ccsdesc[500]{Mathematics of computing~Probabilistic representations}

%%

\maketitle

\section{Introduction}
\label{sec:Introduction}

Discrimination refers to an unjustified difference in treatment toward an individual based on (perceived) membership in a group protected by non-discrimination law \cite{Romei2014MultiSurveyDiscrimination}.
Testing for discrimination is a comparative process centered around the complainant-comparator pair.
As the names would suggest, the \textit{complainant} is the individual instance motivating the discrimination claim while the \textit{comparator} is the individual instance used for validating it.
We use the comparator to answer the hypothetical question behind all discrimination claims: \textit{would the decision have been the same if the complainant had been a member of the non-protected rather than the protected group?}
The comparison between the outcomes of the complainant and its comparator helps establish whether the decision-making process in question is or is not discriminatory.
Most tools for discrimination testing, from the traditional methods like correspondence studies~\cite{Bertrand2004_EmilyAndGreg} and situation testing~\cite{Bendick2007SituationTesting} to the recent algorithmic methods like discrimination discovery~\cite{Ruggieri2010_DMforDD} and counterfactual fairness~\cite{Kusner2017CF}, rely on this pair. 
These tools not only implement the complainant-comparator comparison but also derive the complainant's comparator as a whole based on the relevant process data.

The comparator is central to establishing evidence for discrimination.
Non-discrimination law requires similar, or similarly situated, individuals to be treated similarly regardless of gender, race, or other protected attributes.
It is a requirement shared across countries with only some variation on the list of protected attributes~\cite{Romei2014MultiSurveyDiscrimination}.
The comparator is meant to represent a similar individual to the complainant and allows us to envision an alternative path for the complainant conditional on ``changing'' its protected attribute. 
It requires what we view in practice as \textit{an imaginative act} in which we find (using, for instance, matching models~\cite{DBLP:journals/jiis/QureshiKKRP20}) or create (using generative models~\cite{BlackYF20_FlipTest}) an individual instance representative enough of the complainant's alternative world.
In fact, the comparator, unlike the complainant that motivates it, can be a fictitious individual instance~\cite{DBLP:conf/fat/WeertsXTOP23}.

The choice of comparator, meaning how these discrimination testing tools derive the comparator, matters for three incremental reasons.
First, it embodies a normative statement on what similarity between individuals is and, in turn, how the protected attribute influences individual outcomes within the context of interest \cite{DBLP:conf/fat/WeertsXTOP23}.
Second, it conditions the results of these tools by determining which individual outcomes are to be compared for evaluating the discrimination claim \cite{DBLP:journals/jair/AlvarezR25}.
Third, it acts as a target for the sort of substantive equality between individuals that we wish to achieve as a society through non-discrimination law \cite{Wachter2020BiasPreserving}.
Most tools aim to define the comparator as an observably similar instance to the complainant all but for membership in the protected attribute, such as defining a male candidate with the same work experience as the female complainant in a hiring decision or a white applicant with the same credit score as the non-white complainant in a lending decision.
We refer to this standard operationalization of the comparator as the \textit{ceteris paribus}, or ``with all else being equal,'' (CP) comparator.
Precisely because of these three reasons, debates around the CP comparator predate modern decision flows that rely on machine learning (ML) \cite{Westen1982EmptyEquality, Heckman1998_DetectingDiscrimination, Kohler2018CausalEddie}.
These debates remain relevant as ML researchers tackle the new problem of algorithmic discrimination \cite{Kleinberg2019DiscriminationAgeOfAlgorithms, DBLP:conf/aaai/Ruggieri0PST23, DBLP:journals/ethicsit/AlvarezCEFFFGMPLRSSZR24}.

This work studies the problem of testing for discrimination by revisiting the role of the comparator. 
We do so by, first, embracing the causal modeling nature of discrimination testing and, second, proposing an alternative to the standard CP comparator.
Regarding the first point, given that the comparator answers the ``what would have been if'' question around the complainant, we assert that the comparator is a counterfactual representation of the factual complainant. We further assert that the derivation of the comparator itself constitutes a causal modeling problem and that such problem formulation is the best possible one given the comparative nature of discrimination testing.
Regarding the second point, we propose the \textit{mutatis mutandis}, or ``with the appropriate adjustments being made,'' (MM) comparator. The MM comparator requires that all non-protected attributes constituting the comparator account for the downstream effects of the protected attribute.
Hence, the CP comparator represents an idealized comparison to the complainant while the MM comparator represents an adjusted comparison to the complainant with each based on how we conceive modeling-wise the influence of the protected attribute on all other attributes.
Example~\ref{ex:RunningExample} motivates and illustrates the difference between the two comparators.
Altogether, we view the implementation of the MM comparator as an impactful application for ML models.
Implementing the MM comparator relative to the CP comparator imposes additional complexity (read, a stronger imaginative act) to the problem of discrimination testing that can be better addressed by ML models over more traditional models.
This additional point contrasts with popular works, mainly within the fair ML field \cite{Kohler2018CausalEddie, Hu_facct_sex_20}, that are adamant against treating discrimination testing as a ML problem.

\begin{example}(Tenure at U)
\label{ex:RunningExample}
    \footnote{Example based on~\citet{Morgan2021_UnequalImpactParenthoodAcademia}'s research on the unequal impact of parenthood in academia for female researchers. Social expectations and household dynamics are key factors in creating this difference. Mothers, as the primary caregivers, tend to carry the burden of childbearing within their households, which consumes potential time that could be spent on advancing their academic career by, for instance, writing papers at home. Fathers tend to share less of this burden. See~\citet{Morgan2021_UnequalImpactParenthoodAcademia} for details. Also see~\citet{RomanRodriguez2023} for similar results in couples outside of academia.}
    Let us assume that the university U only considers the number of publications for granting tenure. Clara, a female researcher, was denied tenure for having published only 12 papers and files a discrimination claim against U, becoming the complainant. Suppose that we use Mike, a male researcher who also published 12 papers and was denied tenure by U, as her comparator. Do we deny Clara's discrimination claim? Given what we know about parenthood's impact on academic performance, would our answer change by considering that Clara is a mother? Assuming some quantifiable penalty on the number of publications from motherhood, would we be willing to use Vincent, with 18 publications and recently tenured, as Clara's comparator instead of Mike?
\end{example}

In Example~\ref{ex:RunningExample}, Mike represents the CP comparator (Definition~\ref{def:TheStandardComparator}) while Vincent represents the MM comparator (Definition~\ref{def:MM-Comparator}). 
Depending on which one we use, assuming a literal comparison, we would invalidate or validate, respectively, Clara's discrimination claim. 
Despite its simplicity, Example~\ref{ex:RunningExample} captures the intricacies behind discrimination testing once we contextualize the decision-making process based on information about the protected attribute.
Given what we know from works like~\citet{Morgan2021_UnequalImpactParenthoodAcademia}, in this example, we are implicitly asking which male researcher is more similar to Clara given what we know about gender and its role in this context. 
We believe that it is not an easy question to answer, highlighting the complexity behind modeling the comparator.
With Example~\ref{ex:RunningExample} and, overall, with this work we are not suggesting that the CP comparator is wrong or that the MM comparator is better.
We simply want to position the MM comparator as a viable alternative formulation to the standard CP comparator and, in doing so, stress the importance of the choice of comparator in discrimination testing.
Policies like non-discrimination law and positive actions aimed at addressing systematic inequalities require concrete targets~\cite{Wachter2020BiasPreserving}.
Discrimination testing tools enforce such targets through the comparator.
The CP and MM comparators, thus, represent two distinct targets.
We believe that it is a missed opportunity to rely solely on the CP comparator, especially when we already have ML modeling tools able to implement the more complex MM comparator.
In this work, we focus on one of these ML tools by using structural causal models (SCM) \citet{Pearl09} to derive the MM comparator. 

\paragraph{\textbf{Contributions.}}
Our main contributions are twofold. 
\textit{First}, we demonstrate why the derivation of the comparator is best understood as a causal modeling problem centered on counterfactual reasoning.
This is indeed a divisive view on establishing discrimination but one that should be embraced for developing better discrimination testing tools.
\textit{Second}, with the MM comparator we propose a new approach to the derivation of the comparator and introduce the distinction between CP and MM comparators.
This novel distinction expands the methodological landscape for establishing discrimination.
\textit{Additionally}, our illustrative experiment showcases how this distinction in comparators impacts discrimination testing in practice.
The need for implementing the more imaginative MM comparator clearly positions ML models as promising approaches within discrimination testing.

\paragraph{\textbf{Structure.}} 
In the remainder of this section, we discuss our modeling choice for causality (Section~\ref{sec:Causality}) and relevant related work (Section~\ref{sec:RelevantRW}). 
In Section~\ref{sec:ProblemFormulation}, we argue that deriving the comparator is a causal modeling problem and formulate the CP comparator.
In Section~\ref{sec:MM}, we propose the MM comparator and justify its departure from the CP comparator.
In Section~\ref{sec:Experiments}, using a popular real-world dataset in fair ML, we showcase both comparators by comparing two related discrimination testing tools---situation testing~\cite{Thanh_KnnSituationTesting2011} and counterfactual situation testing~\cite{CST23, DBLP:journals/jair/AlvarezR25}---that implement, respectively, the CP comparator and MM comparator.
In Section~\ref{sec:Survey}, we provide a representative survey of discrimination testing tools.
In Section~\ref{sec:Conclusion}, we conclude this work.

\paragraph{\textbf{Audience, scope, and word choice.}} 
Discrimination testing is an interdisciplinary topic. 
Our work is aimed primarily at ML researchers.
We are interested in the quantitative tools used for deriving \textit{prima facie} evidence for discrimination claims, which constitutes one of many aspects of the legal process for establishing discrimination.\footnote{Technically, all discrimination tools can only test for \textit{prima facie} discrimination \cite{Romei2014MultiSurveyDiscrimination, DBLP:conf/fat/WeertsXTOP23, DBLP:journals/ethicsit/AlvarezCEFFFGMPLRSSZR24}. This distinction, however, is sometimes downplayed in the literature. We come back to this distinction in Section~\ref{sec:ProblemFormulation.Discrimination}.}
Throughout the paper, we focus on \textit{individual discrimination}. 
Namely, we are interested in testing whether an individual instance of a protected group has been discriminated against. The notion of group discrimination instead tackles discrimination against a subset of individuals from a protected group.\footnote{However, \citet{DBLP:conf/fat/Binns20} argues that the apparent conflict between individual and group fairness is an artifact of fairness metrics rather than reflecting different normative principles. Here, we stick with the current fair ML convention.} 
Notably, although often used interchangeably, we make a clear distinction between unfair and discriminatory decisions. Both concepts focus on judging the decision outcomes around a specific attribute; however, discrimination has a stronger meaning than unfairness as it requires such attribute to be under non-discrimination law. Unfairness, in that sense, is more flexible in meaning. All discriminatory decisions are unfair, but not all unfair decisions are discriminatory. Discrimination, in short, can be viewed as unfairness sanctioned by law.
Our focus is mainly on the operationalization of the comparator, which we view as a ML problem that holds for any modern decision flow, regardless of the nature of the decision-maker.
Throughout the paper, we use \textit{generative model} to refer to a model capable of creating synthetic data~\cite{McElreath2018StatisticalRethinking}. 
These are models capable of learning enough about the data generating process of interest to generate instances that are both novel and conceivable under said process, which includes, by definition, approaches like representation learning \cite{Zemel2013LearningFairRepresentations} and SCM \citet{Pearl09}.
Similarly, we use \textit{counterfactuals} to refer to interventional probability distributions obtained through causal reasoning~\cite{Pearl09}, not to be confused with counterfactual explanations from explainable AI~\cite{DBLP:journals/access/StepinACP21,DBLP:journals/datamine/Guidotti24}.\footnote{Counterfactual explanations refer to individual instances similar to the one that is explained but have a different outcome.
To avoid confusion, \citet{DBLP:journals/ai/Miller19} proposes to name them contrastive explanations, and reserve the term counterfactual as understood in the statistical causality literature.} 

\subsection{Auxiliary Causal Knowledge}
\label{sec:Causality}

We formulate causality using \textit{structural causal models} (SCM)~\cite{Pearl09}, a popular framework within causal ML.
SCM are probabilistic graphical models that link (latent) exogenous variables to (observable) endogenous variables via structural equations. 
Each SCM describes a probability distribution and can generate hypothetical distributions via interventions.
Formally, a SCM $\mathcal{M}$ is a tuple $\mathcal{M} = \langle \mathbf{U}, \mathbf{X}, \mathbf{F} \rangle$, where $\mathbf{U} = (U_1, \ldots, U_p)$ are $p$ independent exogenous latent variables, with $\mathbf{U} \sim P(\mathbf{U}) = \prod_{j} P(U_j)$, $\mathbf{X} = (X_1, \ldots, X_p)$ are $p$ endogenous observed variables, and $\mathbf{F} = (f_1, \ldots, f_p)$ are structural equations such that:
\begin{equation}
\label{eq:SCM}
   \text{for $j=1, \dots, p$} 
    \;\;\; 
    X_j := f_j(X_{pa(j)}, U_j).
\end{equation}
Each $f_j$ defines the observed variable $X_j$ on the basis of its causal parents $X_{pa(j)}$ and of the latent variable $U_j$. 
Associated with the SCM $\mathcal{M}$ is the causal graph $\mathcal{G}$, in which each node is a random variable and the directed edges between them a causal effect.
We assume a directed acyclic graph, or DAG, $\mathcal{G}$.
Through $\mathcal{G}$, we represent how the variables relate to each other in terms of cause-effect pairs.
The causal graph $X_1 \rightarrow X_2$, for instance, reads as ``$X_1$ causes $X_2$''.

The SCM $\mathcal{M}$ in \eqref{eq:SCM} acts as a generative model by generating counterfactual distributions that answer counterfactual questions such as the ones asked in discrimination claims. Let $A$ be one of the variables in $\mathbf{X}$ denoting the protected attribute.
We calculate the counterfactual variables $\mathbf{X}^{CF}_{A \leftarrow \alpha}$ resulting from setting $A := \alpha$ through a three-step procedure consisting of abduction, action, and prediction~\cite{Pearl2016_CausalInference}.
\textit{Abduction}: for each prior distribution $P(U_j)$ that describes $U_j$, we compute its posterior distribution given the evidence, or $P(U_j \;| \; \mathbf{X})$. 
\textit{Action}: we intervene on $A$ by changing its structural equation to $A := \alpha$, which gives way to a new SCM $\mathcal{M}'$. 
\textit{Prediction}: we generate the \textit{counterfactual distribution} $P(\mathbf{X}^{CF}_{A \leftarrow \alpha} \; | \; \mathbf{X})$ by propagating the abducted $P(U_j \;| \; \mathbf{X})$ through the revised structural equations in $\mathcal{M}'$.

Importantly, beyond Section~\ref{sec:Experiments}, in which we rely on counterfactual generation to implement the MM comparator, what we discuss in Sections~\ref{sec:ProblemFormulation} and \ref{sec:MM} is not exclusive to SCM.
It is our generative model of choice, mainly because previous works on discrimination testing \cite{Kusner2017CF, DBLP:conf/nips/KilbertusRPHJS17, CST23, DBLP:journals/jair/AlvarezR25} use it.
Our goal is not to prioritize one generative modeling approach over the others, but to focus on one generative model approach to show how the others could be useful for the problem of discrimination testing under the MM comparator.

\subsection{Related Work}
\label{sec:RelevantRW}

The literature on discrimination testing is vast, multidisciplinary, and regional. 
We refer to~\citet{DBLP:journals/tosem/ChenZHHS24} for a recent survey on testing the fairness of ML models, and to~\citet{DBLP:conf/fat/HutchinsonM19} and \citet{Romei2014MultiSurveyDiscrimination} for surveys with a multidisciplinary perspective. 
For a general overview on bias and fairness in AI, see~\citet{DBLP:journals/ethicsit/AlvarezCEFFFGMPLRSSZR24,DBLP:journals/widm/NtoutsiFGINVRTP20} and \citet{DBLP:journals/csur/MehrabiMSLG21}. 

Here, we wish to position our work with respect to three papers that are directly applicable to the modeling problem of revisiting the comparator in discrimination testing. 
To the best of our knowledge, we are the first to explicitly tackle this problem. 
Later in Section~\ref{sec:Survey} we provide a selective survey of discrimination testing tools, with a focus on the comparator chosen by those tools.
We do so as we first want to introduce the CP and MM comparators in Sections~\ref{sec:ProblemFormulation} and \ref{sec:MM}, and illustrate their impact on a discrimination testing problem in Section~\ref{sec:Experiments}.

First,~\citet{Kohler2018CausalEddie} argues that discrimination testing focuses on answering counterfactual questions about the complainant. 
The work refers to the current approach as the \textit{causal counterfactual model of discrimination}, and states that it is flawed due to its treatment of protected attributes. In particular, using a constructivist approach to race, the work argues that comparing similar individuals that only differ on skin color fails to capture how race affects all other aspects of these individuals' lives. We draw considerably from~\citet{Kohler2018CausalEddie} as it represents the first critique of the CP comparator. However, different from~\citet{Kohler2018CausalEddie}, we propose an alternative to the CP comparator in the form of the MM comparator and embrace the role of counterfactual reasoning within discrimination testing. This last point is important as Kohler-Hausmann's view (see, for instance,~\citet{Hu_facct_sex_20}) is largely against causal modeling for discrimination testing and is suspicious toward attempts to measure the so-called causal effect of race. Kohler-Hausmann fails, though, to provide an alternative to the current model. 
We disagree with \citet{Kohler2018CausalEddie}'s position toward causal modeling, but agree that we can and should do better than the CP comparator. 
The MM comparator is our response to~\citet{Kohler2018CausalEddie}.

Second,~\citet{Wachter2020BiasPreserving} states that non-discrimination law aims for \textit{substantive equality}, meaning it should aim at changing the status quo of the societal context of interest.\footnote{The argument is specific to the European Union. Although it seems not to be the case for US non-discrimination law~\cite{Barocas2016_BigDataImpact}, the goal of substantive equality is implicit in many of the fair ML definitions used by US researchers as argued by~\citet{Wachter2020BiasPreserving}.} 
Under this premise, the work classifies fair ML methods into bias preserving and bias transforming based on whether they assume (the former) or not (the latter) a neutral status quo. Although~\citet{Wachter2020BiasPreserving} makes a strong case for ML methods as societal shapers, the work lacks clarity on how these methods are to be implemented, especially with respect to non-discrimination law. We tackle this gap by formalizing the MM comparator. Our reading of~\citet{Wachter2020BiasPreserving} is that non-discrimination law must not aim at reaching equality of outcomes only, but also at imposing equality of opportunities in a systematic way. Such a step requires envisioning an ideal target, which needs to be generated for the current context to aim at. The MM comparator represents the goal of substantive equality in the context of discrimination testing.

Third,~\citet{CST23} introduces the CP and MM classification but briefly and without the necessary discussion.
\footnote{The same criticism applies to the journal version: \citet{DBLP:journals/jair/AlvarezR25}. We do this on purpose. The CP versus MM discussion merits its own work as it applies to all discrimination testing tools.}
This is because the focus of the paper is on formulating the discrimination testing method of counterfactual situation testing, which is a possible implementation of the MM comparator.
Drawing from~\citet{Kohler2018CausalEddie}'s discussion on the causal counterfactual model of discrimination, the work states that discrimination testing tools can be broadly classified into comparators that do and do not consider the downstream effects of the protected attribute on all other attributes. 
The work offers a valuable first step to the goal of revisiting the comparator, but falls short in formalizing the claims on the comparator. 
The work neither defines these kinds of comparators nor surveys the quantitative tools to illustrate them. 
On top of addressing these shortcomings from~\citet{CST23}, we discuss the role of counterfactual reasoning in discrimination testing, which the work takes for granted.

We note that the causal treatment of discrimination using SCM is a common practice. 
See~\citet{DBLP:journals/jlap/MakhloufZP24} for a survey. 
\citet{DBLP:conf/nips/KilbertusRPHJS17} are the first to formulate problems like direct, indirect, and proxy discrimination using SCM, while~\citet{DBLP:journals/ftml/PleckoB24} try to summarize discrimination within a general causal fairness model using SCM. 
Works like these are relevant, but lack interdisciplinary depth. For instance, both papers fail to discuss non-discrimination law and how exactly the modeling tools based on their formulations can be or need to be used as evidence. 
Their treatment of the subject is detached from the real-world application. 
These papers do not discuss the comparator. 
To be clear, such papers have a different scope---mainly to formulate discrimination using SCM---but, because of such approach, fail to address the role of causality in discrimination testing.
Such a discussion is more explicit in the social sciences, in which causal models are also used for testing discrimination. See, for instance, \citet{Heckman1998_DetectingDiscrimination}. 
Our work complements these and similar works by revisiting the comparator. 

%
% EOS
%

\section{The Comparator as a Causal Modeling Problem}
\label{sec:ProblemFormulation}

In this section, we argue that testing for discrimination amounts to a two-step modeling problem centered on deriving the comparator $i'$ and comparing it to the discrimination claim's complainant $i$. 
We advocate for the derivation of the comparator, motivating the need for $i'$. 
We also advocate that the derivation of the comparator is a causal modeling problem not only because we use $i'$ to demonstrate whether or not the decision is caused by the protected attribute, but also because $i'$ is a counterfactual representation of the factual $i$.

\subsection{The Setup}
\label{sec:ProblemFormulation.SetUp}

Non-discrimination law mandates that a decision-maker, be it algorithmic or human, treats similar individuals similarly regardless of the protected attribute \cite{Kleinberg2019DiscriminationAgeOfAlgorithms}.
Discrimination testing tools uncover discrimination by finding and comparing similar individuals to isolate the effect of the protected attribute on the decision outcome \cite{Romei2014MultiSurveyDiscrimination}.
Formally, letting $A$ represent the protected attribute, $\mathbf{X}$ the set of non-protected attributes, and $Y$ the decision outcome, we write this two-step modeling problem into Definitions \ref{def:TheStandardComparator} and \ref{def:DiscriminationTesting}, respectively.

\begin{definition}(The CP Comparator)
\label{def:TheStandardComparator}
    Let the real individual profile $\langle \mathbf{x}_i, a_i, y_i \rangle$ represent the complainant $i$ on which the discrimination claim is based. 
    We define the \textit{ceteris paribus} (CP) comparator $i'$ as a real or fictitious individual profile $\langle \mathbf{x}_{i'}, a_{i'}, y_{i'} \rangle$ such that:
    \begin{equation}
    \label{eq:TheStandardComparator}
        d\big(\mathbf{x}_i, \mathbf{x}_{i'}\big) \leq \epsilon \;\;\; \text{and} \;\;\; a_i \neq a_{i'}
    \end{equation}
    where $d(\cdot, \cdot)$ is a distance function and $\epsilon \in \mathbb{R}^+$ a threshold.\footnote{The issue of engineering fairness is challenging \citep{DBLP:journals/ethicsit/Scantamburlo21}. We believe that the definition of the distance function $d(\cdot, \cdot)$ is not only a technical choice (see the survey by \citet{DBLP:journals/pr/BlancoMalloMRB23}), but is also specific to the application domain. For instance, in job screening, $d(\cdot, \cdot)$ can result from a weighting of the skills of job candidates: how to set the weights is a non-trivial issue, which requires an agreement among multiple stakeholders. For this reason, in this paper, we make no assumption/preference about the definition of $d(\cdot, \cdot)$.} The closer $d(\cdot,\cdot)$ is to $0$, the more similar are $\mathbf{x}_i$ and $\mathbf{x}_{i'}$.
\end{definition}

We emphasize that it is possible to provide a hypothetical comparator \cite{DBLP:conf/fat/WeertsXTOP23}, meaning that we may find or generate an $i'$ similar under $d$ to $i$.
The comparator derived from Equation~\eqref{eq:TheStandardComparator} is the standard comparator in discrimination testing \cite{Heckman1998_DetectingDiscrimination, Kohler2018CausalEddie}.\footnote{There is a conception of discrimination testing that is non-comparative, though it is not widely used \cite{DBLP:conf/fat/Binns20}.}
We refer to it as the \textit{ceteris paribus}, or ``with all else being equal,'' (CP) comparator as, ideally, we expect to find an $i'$ profile all equal but for the protected attribute to the $i$ profile when $\epsilon=0$.
Intuitively, it operationalizes the basic axiom underpinning non-discrimination law---that of treating similar individuals similarly---especially when the decision-maker $f$ is, realistically, only allowed to use the set of non-protected attributes when making a decision, meaning $Y = f(\mathbf{X})$. 
In Section~\ref{sec:MM}, we come back to the CP comparator and revisit it explicitly.

\begin{definition}(Difference in Outcomes)
\label{def:DiscriminationTesting}
    Given the individual profiles $\langle \mathbf{x}_i, a_i, y_i \rangle$ and $\langle \mathbf{x}_{i'}, a_{i'}, y_{i'} \rangle$, we test the discrimination claim by focusing on the difference in the same decision outcome:
    \begin{equation}
    \label{eq:DiscriminationTesting}
        \delta 
        % = P(Y=y_{i'}) - P(Y=y_i) 
        = P(Y=y_{i'} \, | \, \mathbf{X}=\mathbf{x}_{i'}, A = a_{i'}) - P(Y=y_i \, | \, \mathbf{X}=\mathbf{x}_{i}, A = a_{i}) 
    \end{equation}
    where $\delta$ may, for example, be greater than or equal to 0 or need to follow a known distribution depending on the legal context.
    Notably, we focus on the same decision between the complainant and comparator, meaning $y_{i'}=y_i$.
\end{definition}

The difference in outcomes \eqref{eq:DiscriminationTesting} is not based on a literal comparison, meaning $y_{i'} - y_i$, but on a difference in probabilities. 
It implies the need to construct some measure of uncertainty (hence, why we write $\delta$ in terms of probabilities) around the decision outcome for each individual profile.
Formally, $\delta$ is the difference between two point estimates that requires the use of a statistical estimator. 
To obtain it, therefore, we need either multiple observations centered on these two profiles or strong assumptions about the behavior of these two profiles.
Legal experts require or, at least, expect this kind of comparison \cite{EU2018_NonDiscriminationLaw}. We believe that this practice should not come as a surprise to the reader. For instance, if it takes more than one throw to check whether a coin is fair, why should it not be the same for a serious accusation like discrimination? Intuitively, we wish to control for chance from (or rule out uncertainty in) the decision process in question to have some certainty on the pattern we are testing for. This is not possible under a literal comparison. 

\textit{These two definitions together describe the general setup implicit in all discrimination testing tools}.
Given our focus on revisiting the standard CP comparator, in this work we focus on Definition~\ref{def:TheStandardComparator}.
Clearly, whatever we derive as the comparator following Equation~\eqref{eq:TheStandardComparator} conditions what outcomes are compared in Equation~\eqref{eq:DiscriminationTesting} and, in turn, what we test for and uncover as discrimination when using a discrimination testing tool.

\subsection{On (Establishing) Discrimination}
\label{sec:ProblemFormulation.Discrimination}

The comparator is intrinsic to how we conceive discrimination, and this conception is comparative \cite{Westen1982EmptyEquality, DBLP:conf/fat/WeertsXTOP23, Kohler2018CausalEddie}.
Now, overall, discrimination can be both a familiar and non-trivial concept.
We draw a distinction between describing versus proving something as discriminatory, which helps understand discrimination as a social phenomenon (the former) and as a modeling problem (the latter).
Let us start with describing discrimination. 
Quoting from \citet{DBLP:conf/fat/WeertsXTOP23}, who base their own definition on \citet{Lippert2006BadnessOfDiscrimination}: 
\begin{displayquote}
    Discrimination can generally be characterized by the morally objectionable practice of subjecting a person (or group of persons) to a treatment in some social dimension that, for no good reason, is disadvantageous compared to the treatment awarded to other persons who are in a similar situation, but who belong to another socially salient group.
\end{displayquote}
A group is considered to be socially salient ``if perceived membership of it is important to the structure of social interactions across a wide range of social contexts''~\cite{DBLP:conf/fat/WeertsXTOP23}.
Defining the socially salient group can only be done by agreeing on a specific, shared social context. 
Arriving at this point by recognizing certain groups over others under non-discrimination law captures the social phenomenon.

There is a clear comparative element to this view on discrimination. 
We are not interested in the difference in treatment between any two individuals from different socially salient groups, but in the difference in treatment between two similar (or similarly situated) individuals from different socially salient groups.
Discrimination occurs when similar (or similarly situated) individuals that differ in membership in a socially salient group are treated differently. 
Under this setting, discrimination becomes the opposite of equality \cite{DBLP:conf/fat/WeertsXTOP23}.
Also under this setting, the comparator becomes a manifestation of what similar (or similarly situated) individuals look like.

Let us now address proving discrimination. 
We consider the European Union (EU) context because, one, most fair ML works consider only the US context and, second, the EU---with the AI Act and GDPR---currently leads on the regulation front for algorithms \cite{DBLP:journals/ethicsit/AlvarezCEFFFGMPLRSSZR24}.
In doing so, we want to show the complex pipeline behind establishing discrimination.
Although the discussion below concerns EU non-discrimination law, the modeling problem formalized in Section~\ref{sec:ProblemFormulation.SetUp} is shared among most Western conceptions of justice \cite{Westen1982EmptyEquality} and it is, thus, a robust problem to the legal context.

There are two forms of discrimination conceived by EU non-discrimination law: direct and indirect discrimination.
Under direct discrimination, the decision-maker uses the protected attribute for the decision, while under indirect discrimination the decision-maker uses a supposedly neutral attribute that is informative of the protected attribute.\footnote{In US non-discrimination law, direct discrimination is referred to as disparate treatment and indirect discrimination as disparate impact \cite{Barocas2016_BigDataImpact}. The EU context almost fully translates onto the US context, especially when it comes to direct discrimination and disparate treatment. For our purposes, the goal of substantive equality over formal equality is specific to the EU \cite{Wachter2020BiasPreserving}. We come back to this point in Section~\ref{sec:MM}, but, regarding the overall problem of the comparator, we can generalize across the two legal contexts. The same seems to hold for UK non-discrimination law \cite{DBLP:journals/corr/abs-2407-00400}.}
There are four elements in a discrimination claim, which we list below in order \cite{DBLP:conf/fat/WeertsXTOP23}. 
We add in italics the implications to modern decision flows.
\begin{itemize}
    \item \textbf{``On grounds of''...} 
    We need to determine whether the claim falls under direct or indirect discrimination, which will depend on whether the decision is taken based on the protected attribute or not. 
    \textit{This element has been interpreted in terms of whether or not the protected attribute is an input to the ML model \cite{Hacker2018TeachingFairness}: currently, models are not allowed to use protected attributes. Recent works argue that if the model does not use the protected attribute, but is able to infer it (read, proxy discrimination \cite{Tschantz2022_ProxyDisc}), then it should still fall under direct discrimination \cite{Adams2022DirectlyDiscriminatoryAl}.}
    \item \textbf{...``a protected characteristic.''} 
    The claim must be based on an attribute protected by non-discrimination law, such as gender or race. These are attributes that summarize historical processes as they were once used, or are still in use, to not only classify individuals but to purposely shape their social and economic opportunities \cite{Sen2016_RaceABundle, Bonilla1997_RethinkingRace, TheTroubleWithDisparaty}. Non-discrimination law acknowledges the historical injustices toward members of these groups by encoding them as ``protected''. That is why a discrimination claim is specific to non-discrimination law, meaning it cannot be applied to any characteristic of an individual, only to a protected one.
    \textit{This element, on top of the current discussions on intersectional versus multiple discrimination \cite{Xenidis2020_TunningEULaw}, is ongoing as there is a fear that models are creating new protected attributes.}
    \item \textbf{... where there is evidence for ``less favorable treatment''...}
    The complainant, to establish a case, needs to show \textit{prima facie} evidence, meaning ``sufficient evidence for a rebuttable presumption of discrimination to be established by the judge'' \cite{DBLP:conf/fat/WeertsXTOP23}.
    If \textit{prima facie} discrimination is established, the burden of proof falls on the defendant.
    \textit{Under a ML model, evidence itself becomes standardized. In principle, we must focus on the input-output pair of the model in question.}
    \item \textbf{...unless there is an ``objective justification.''}
    Direct discrimination cannot be justified in principle. Indirect discrimination, instead, can be justified as long as it has a legitimate goal and passes the proportionality test. 
    As \citet{DBLP:conf/fat/WeertsXTOP23} point out, neither the law provides concrete guidelines for the proportionality test nor can it be settled in advance. 
    Further, the ``objective justifications'' are settled on a case-by-case basis.
    \textit{This element shows that we can train a ML model that could be used across multiple cases but to judge its fairness (for the purpose of testing discrimination) will depend on each case \cite{Wachter2021FairnessCannotBeAutomated}.}
\end{itemize}
The third element is the one that concerns the comparator as it requires the use of the discrimination testing tools for obtaining the \textit{prima facie} evidence. 
Proving discrimination, similar to describing discrimination, has a clear comparative element; here, though, it also has a modeling angle because the goal of obtaining evidence is explicit.
The idea is still to compare similar (or similarly situated) individuals, though explicitly by defining $d$ in Definition~\ref{def:TheStandardComparator}, materializing what similarity between individuals is within our context. 

Defining similarity is precisely the main challenge inherent to modeling the comparator $i'$ with respect to the complainant $i$.
The social context that describes discrimination can be and often is taken for granted when modeling the comparator as it is well understood. 
In Example~\ref{ex:RunningExample}, for example, we did not need to motivate females and males as socially salient groups. 
The same occurs with most fair ML papers that simply list protected attributes like race and gender to which the proposed methods apply \cite{DBLP:conf/aaai/Ruggieri0PST23}.
However, a well-understood context still does not release us from making normative statements that are socially charged when we define similarity between individuals through $d$.
We cannot escape defining $d$ either as only through $d$ we derive the comparator $i'$ for the complainant $i$.
In principle, as long as some suitable $d$ exists, the modeling problem of proving discrimination appears simple and even intuitive.

The simplicity, however, breaks down precisely because we must first define similarity between individuals to then obtain similar individuals and, in turn, prove discrimination.
In \textit{The Empty Idea of Equality}, \citet{Westen1982EmptyEquality} points out this limitation of non-discrimination law by arguing that equality is a circular concept: to argue for equality we first must define what it means to be equal. 
It follows, thus, that \textit{similarity is a circular concept}.
The circularity of similarity matters because, although a discrimination tool's main goal is to find evidence, which on paper might seem an objective goal, any discrimination testing tool---be it implicitly or explicitly---needs to implement a normative statement on similarity in the form of some $d$ in order to be operational. 
It is a difficulty inherent to deriving the comparator, be it the standard one or not.
Hence, none of these tools can be sold as unbiased in practice.
 
Given that $d$ is never fully objective, a valid question for concluding this section is whether similarity can ever be defined without mathematical formalism. 
It is possible, but not the common approach, especially when it comes to testing discrimination.
In philosophical works on discrimination (see, e.g., \citet{Loi2023_CounterfactualFairnessDisc}), the question of similarity is addressed through argumentation.
The main drawback we see with such an approach is the lack of scalability. 
Granted that mathematical formalism may give a false sense of objectivism relative to argumentation \cite{Kohler2018CausalEddie, Hu_facct_sex_20}, the fact remains that establishing discrimination requires evidence. 
Simply put, if we need to find or generate other similar individuals to the complainant in order to build our evidence, then using $d$ is more convenient than arguing the similarity of multiple individuals to the complainant on a case-by-case basis.

\subsection{The Counterfactual Model of Discrimination}
\label{sec:ProblemFormulation.CausalModeling}

Discrimination is often conceived as a causal claim on the (in)direct effect of the protected attribute $A$ on the decision outcome of interest $Y$ \cite{Heckman1998_DetectingDiscrimination}.
Its causal underpinning, which is motivated by non-discrimination law's own definition of discrimination \cite{DBLP:conf/fat/WeertsXTOP23}, has long motivated a range of methods for testing discrimination based on \textit{counterfactual reasoning}.
These methods operationalize the scenario in which we are able to manipulate the protected attribute of the individual(s) making the claim, imagine the ``what would have been if'' (or the counterfactual) outcome, and compare it to the ``what is'' (or the factual) outcome to isolate the causal effect of $A$ on $Y$.
\citet{Kohler2018CausalEddie} refers to this practice as the \textit{counterfactual model of discrimination}.\footnote{\citet{Kohler2018CausalEddie} actually refers to it as the counterfactual causal models of discrimination, which is redundant as counterfactuals are causal \cite{Woodward2005MakingThigsHappen}. Hence, we drop the ``causal.''}
Using SCM, we can state the discrimination claim as $A \rightarrow Y$ (if direct) or as $A \rightarrow \mathbf{X} \rightarrow Y$ (if indirect). 
The goal of any discrimination testing tool is validating such claim.

What we want to emphasize about the counterfactual model of discrimination is not that it relies on counterfactual reasoning to answer causal claims, but that, because of this reliance, the comparator $i'$ is a counterfactual representation of the complainant $i$.
Viewing $i'$ as a counterfactual representation of the factual $i$ ultimately positions the comparator as a causal modeling problem. 
This additional emphasis from our side is important when we consider that $i'$ can be either a real or fictitious profile, making it hypothetical in the fullest sense. 
Further, that the comparator $i'$ can be viewed as a thought experiment is not exploited---both conceptually and experimentally---under the CP comparator. 
We come back to this point in Section~\ref{sec:MM}. 
This work diverges from \citet{Kohler2018CausalEddie} precisely because of this emphasis, viewing the dominant counterfactual model of discrimination as an opportunity suited for fair ML.

Why counterfactuals?
We argue, one, because discrimination represents a pattern and, two, because the discrimination claim requires an answer at the individual level. 
Let us unpack each point. 
First, why would discrimination, being a pattern, amount to a causal relation? 
Broadly, the presence of a sufficiently robust pattern suggests a causal relationship, as causality is seen as an invariant force acting on the world.
%\footnote{
Such a view formally amounts to an interventionist account of causality (see \citet{Woodward2005MakingThigsHappen} for details), which is the basis for both SCM \cite{Pearl09} and Potential Outcomes \cite{Angrist2008MostlyHarmless}, meaning the two most popular causal modeling frameworks.
%}
The presence of a causal pattern also suggests, implicitly, that there is little room for chance to explain what we observe.
Discrimination claims are often characterized in those terms, meaning an unfair decision that persists over time toward a certain group of individuals. Because of this persistence, we are (somewhat) certain that the unfair treatment comes from the influence of the protected attribute. 
Intuitively, a discriminatory decision-making process has (un)consciously set up a pattern (i.e., $A \rightarrow Y$ or $A \rightarrow \mathbf{X} \rightarrow Y$) that causes the unfair treatment of protected individuals.
If the goal is to detect such pattern(s) and we agree on the premise that discrimination represents a causal pattern, then counterfactual reasoning is the main approach to do so.
There is a long tradition in causality, especially within SCM, to formulate what constitutes a cause in terms of counterfactual reasoning \cite{Pearl09, Halpern2016_ActualCausality, HalpernPearl2001_PartI, HalpernPearl2001_PartII}: if intervening $A$ induces a counterfactual outcome $Y^{CF}$ different from the factual outcome $Y$, then $A$ must be a cause of $Y$.
%\footnote{
These definitions are of course much more formal---for instance, in terms of probability distributions---but the phrase captures the main idea behind all of them.
%}
Under this setting, to test for the existence of such a pattern would imply examining the impact of $A$ on the decision-making process by changing $A$ itself. 
If there is such a pattern, then observing a change in outcome due to having changed only $A$ (and controlling for everything else) would, first, confirm the existence of the pattern, and, second, confirm the role of $A$ within it.

Second, why would counterfactual reasoning and, in turn, the comparator, answer individual-level queries? 
Although it is trivial to point out, no individual $i'$ can be more similar to another individual $i$ than that same individual $i$.
When testing for discrimination, we are trying to imagine what that same individual would have been like in another life. 
This is the sort of mental exercise we carry out when asking ``what would have been of Clara had she been male?'' in Example~\ref{ex:RunningExample}.
It is a hypothetical question that, regardless, we aim and, to an extent, need to answer with discrimination testing.
We cannot observe simultaneously, for any individual $i$, both his or her ``what is,'' or factual, and ``what would have been if,'' or counterfactual, outcomes. 
We are only certain of the former, factual outcome; and the best we can do is to derive another individual $i'$ that represents (or approximates) the latter, counterfactual outcome of $i$.
Defining similarity between individuals through $d$ to obtain such $i'$ is an attempt at controlling for all factors that might influence an outcome and being able to answer whether one of those factors, i.e., $A$, is a cause of $Y$.
We stress that this line of reasoning is implicit to all causal and non-causal discrimination testing tools using the counterfactual model of discrimination \cite{Kohler2018CausalEddie}. 
Under the counterfactual model of discrimination,
similarity is not only a statement between two individuals $i$ and $i'$, but it is also a statement on the alternative paths attributed to an individual $i$ through the lived experiences of another (real or fictitious) individual $i'$.
Similarity, in turn, becomes also a statement on this factual world and some hypothetical counterfactual alternative.

%
% EOS
%

\section{Mutatis Mutandis}
\label{sec:MM}

In this section, we revisit the CP comparator of Definition~\ref{def:TheStandardComparator}, which is the standard in discrimination testing, by introducing the \textit{mutatis mutandis}, or ``with the appropriate adjustments being made,'' (MM) comparator.
We first define the MM comparator below.
We then motivate it and discuss how it extends the problem of discrimination testing.

\begin{definition}(The MM Comparator)
\label{def:MM-Comparator} 
    Let $\tilde{d}(\cdot, \cdot)$ be a protected-aware distance function that, unlike $d(\cdot, \cdot)$ in Equation~\eqref{eq:TheStandardComparator}, considers both $\mathbf{X}$ and $A$ as inputs, capturing the potential influence of $A$ on $\mathbf{X}$.
    For the complainant $i$ with profile $\langle \mathbf{x}_i, a_i, y_i \rangle$, we define a \textit{mutatis mutandis} (MM) comparator $i'$ as an individual profile $\langle \mathbf{x}_{i'}, a_{i'}, y_{i'} \rangle$ such that:
    \begin{equation}
    \label{eq:MMComparator}
        \tilde{d} ( (\mathbf{x}_i, a_i), (\mathbf{x}_{i'}, a_{i'}) ) \leq \epsilon \;\;\; \text{and} \;\;\; a_i \neq a_{i'}
    \end{equation}
    where $\epsilon \in \mathbb{R}^+$ is a threshold. Recall that $i$ must be a real profile; $i'$ instead can be real or fictitious. 
\end{definition}

Under $\tilde{d}(\cdot, \cdot)$, two instances with different non-protected attributes, $\mathbf{x}_i \neq \mathbf{x}_{i'}$, can be considered similar given information, respectively, about $a_i$ and $a_{i'}$.
The fact that Definition~\ref{def:MM-Comparator} allows for this ``dissimilarity'' between comparator and complainant motivates the MM denomination.
The general idea is that under $\tilde{d}(\cdot, \cdot)$ we should be able, if needed, to account for the influence of $A$ on $\mathbf{X}$ when deriving the $i'$ relative to $i$.
In other words, we ``adjust for what needs to be adjusted'' as a consequence of ``changing'' $A$, allowing us to derive an individual profile less conservative than the one conceived under $d(\cdot, \cdot)$ in Definition~\ref{def:TheStandardComparator}. 
We motivate this argument in the next sub-sections.

Given the meaning behind $\tilde{d}(\cdot, \cdot)$, note that
when $a_i=a_{i'}$, the MM and CP comparators are the same.
As we move within the same group in $A$, we assume that the influence of $A=a$ on $\mathbf{X}$ is shared among all $i$ individuals with $A=a_i$. 
Implicitly, we assume that defining similarity between individuals of the same group in $A$ requires no additional considerations due to shared experience.
Additionally, when $\mathbf{X}$ and $A$ are independent, the MM and CP comparators are the same even if $a_i \neq a_{i'}$ as there are no downstream effects of $A$ on $\mathbf{X}$ to be accounted for.

\subsection{Against Idealized Comparisons}
\label{sec:MM.IdealComparison}

Let us focus on the standard CP comparator.
We argued already that any notion of similarity between individuals implemented through the distance function $d(\cdot, \cdot)$ in Equation~\eqref{eq:TheStandardComparator} is a normative statement. 
We also argued that, in turn, the comparator derived under $d(\cdot, \cdot)$ represents the normative statement. 
This is true for the standard comparator as for any other formulation of the comparator, such as the MM comparator.
Now, there are two characteristics specific to the CP comparator worth stressing. 
First, in Definition~\ref{def:TheStandardComparator}, similarity between individuals comes down to a comparison on the non-protected attributes $\mathbf{X}$ between $i'$ and $i$. The distance function $d(\cdot, \cdot)$ is oblivious to (or unaware of) the protected attribute $A$. This holds whether $A$ and $\mathbf{X}$ are associated or not as $d(\cdot, \cdot)$ does not take it into consideration.
Second, it follows that the counterfactual representation of the complainant $i$ in the form of the comparator $i'$ only varies, ideally, in terms of $A$.
In other words, nothing would have changed for $i$ in $\mathbf{X}$ had she or he had been from the non-protected group.
Because of these two characteristics, we say that the standard CP comparator represents an \textit{idealized comparison} between $i'$ and $i$. 

Is the idealized comparison problematic, though? 
The answer to this question comes down to having a stance on similarity between individuals, and how it should be operationalized through the $d(\cdot, \cdot)$ in Definition~\ref{def:TheStandardComparator}.
We view it as open to discussion.
In this work, we do not have a strong stance against the CP comparator as it can be useful, for example, in uncovering direct discrimination or when an alternative comparator is not possible.
We do, however, view the discussion against it as necessary for discrimination testing. 
This discussion motivated the MM comparator.

Pivotal in criticizing the CP comparator and the idealized comparison it represents is \citet{Kohler2018CausalEddie}.
Under the premise that the comparator $i'$ in Definition~\ref{def:TheStandardComparator} represents a counterfactual representation of the complainant $i$, \citet{Kohler2018CausalEddie} criticizes the counterfactual model of discrimination for ignoring how membership in a protected group conditions the material outcomes of individuals and, thus, limits any meaningful counterfactual analysis between protected and non-protected individuals.
\citet{Kohler2018CausalEddie}, in particular, argues that protected attributes, like race and gender, are social constructs, which is a view widely held by social scientists \cite{Bonilla1997_RethinkingRace, Sen2016_RaceABundle, Hanna2020_CriticalRace}. 
What it means is that these social categories were once (or still are) used not only to classify individuals but to also delimit the opportunities available to them \cite{Mallon2007SocialConstruction}.
According to \citet{Kohler2018CausalEddie},
it is inconceivable to picture a counterfactual white version of a black individual as stated in Definition~\ref{def:TheStandardComparator}.
Under such definition, the ideal comparator $i'$ will have a set of non-protected attributes identical to those of $i$, or $d(\mathbf{x}_i, \mathbf{x}_{i'})=0$.
By envisioning a counterfactual world where $i$ can be exactly as he or she is today, in terms of non-protected attributes, while belonging to the non-protected group,
this mental exercise requires from us to accept that the protected individual $i$ would have arrived at the same point had he or she been a non-protected individual $i'$, which reduces the pervasive influence of $A$.

Following \citet{Kohler2018CausalEddie}'s constructivist argument, Definition~\ref{def:TheStandardComparator} envisions a scenario where we can manipulate $A$ and expect $\mathbf{X}$ to remain unchanged, failing to capture the role of $A$ and raising the question around what exactly are the discrimination testing tools assessing for. 
As she writes in the case of race:
\begin{displayquote}
    The problem with identifying discrimination with the treatment effect of race is that it misrepresents what race is and how it produces effects in the world, and concomitantly, what makes discrimination of race a moral wrong. [...] [I]f the signifiers of racial categories fundamentally structure the interpretation and relevance of other characteristics or traits of the unit, then it is a mistake to talk about identical units that differ only by raced statues.
\end{displayquote}
\citet{Kohler2018CausalEddie} implies that the comparator $i'$ derived in Definition~\ref{def:TheStandardComparator} is wrong and, in turn, the tools for testing discrimination built using such comparator are also wrong.\footnote{\citet{Kohler2018CausalEddie} later extends the above argument for gender \cite{Hu_facct_sex_20}, covering the two main protected attributes.}
We agree partially with her.
First, we do not see all discrimination testing that uses the CP comparator as wrong as it might be the only viable option in many cases.
Second, we do not picture a better modeling framework for discrimination testing than the counterfactual model of discrimination. 
It is not ideal, but it can also improve if we are willing to revisit the derivation of the comparator.
In Appendix~\ref{App:Disparity}, we discuss additional critical work toward discrimination testing from a non-legal angle.

Arguments against discrimination testing are not new. 
Similarly, \citet{Wachter2020BiasPreserving}, by classifying fair ML methods as bias preserving or bias transforming depending on how they define similarity, recently pointed out the limitations of using the conservative CP comparator.
These arguments, however, are largely absent from the discussion around discrimination testing tools.
There, the focus is always on tool development, which is to be expected. 
However, as we already argued, the circularity of similarity requires deeper consideration from the tool developers.

\subsection{A New Kind of Comparator}
\label{sec:MM.SubEquality}

Once we allow for a more complex view on the protected attribute $A$ and its influence on the non-protected attributes $\mathbf{X}$ by, for instance, viewing $A$ as a social construct \cite{Kohler2018CausalEddie} or acknowledging $A$ as a source of inequality \cite{TheTroubleWithDisparaty}, the CP comparator appears insufficient.
Under this view, there is the need to clarify the relationship between $A$ and $\mathbf{X}$, and whether it is in our interest to account for it when testing for discrimination.
The MM comparator, based on $\tilde{d}(\cdot, \cdot)$ in Equation~\eqref{eq:MMComparator}, is meant to clarify and account for the relationship between $A$ and $\mathbf{X}$, embodying what is referred to as ``fairness given the difference'' \cite{CST23}.
We believe that the MM comparator represents a new kind of comparator, aimed at not only proving discrimination but at shaping the social context behind it.

The MM comparator represents an ideal target level of equality instead of the formal one captured by the CP comparator due to its idealized comparison. 
Conceptually, if $\tilde{d}(\cdot, \cdot)$ accounts for the influence of $A$ on $\mathbf{X}$, then by comparing $i'$ to $i$ when $\mathbf{x}_{i'} \neq \mathbf{x}_i$ and $a_{i'} \neq a_i$ we are making a normative statement claiming that these two observably dissimilar individuals should be similar.
Under counterfactual reasoning, since the counterfactual question is around $A$, the comparator $i'$ derived under Definition~\ref{def:MM-Comparator} represents where the complainant would have been if free from the effect of being part of the protected group.
The CP comparator does not allow for this possibility as it only envisions a comparator $i'$ all equal but for $A$ to the complainant $i$.
That is why in Example~\ref{ex:RunningExample}, Mike is the CP comparator while Vincent is the MM comparator. Vincent, with his higher number of publications, represents a Clara that would have carried out her academic career without the social expectations of being the main caregiver of her household.

The MM comparator acts as a target because we, essentially, need to generate it. 
In Example~\ref{ex:RunningExample}, for example, why should we consider Vincent when other profiles could also meet the MM comparator definition?
In other words, we need to have a modeling procedure that motivates the choice of the MM comparator.
In practice, it means estimating the effect of $A$ on $\mathbf{X}$, and adjusting the latter accordingly.
We discuss this last point in the next section.
Under this ``modeling'' interpretation, the proposed MM comparator might seem abstract. 
In fact, why complicate the derivation of the comparator under the MM definition when we already have the CP definition?
A reason is that, as argued by \citet{Wachter2020BiasPreserving}, non-discrimination law, at least in the EU, aims at substantive equality.
In practical terms, it means that when testing for discrimination, we are not interested in whether the treatment was the same for similar individuals, but whether the setting (\citet{Wachter2020BiasPreserving} refers to it as the ``status quo'') is suitable for similar individuals to be treated similarly over time. 
This point is similar to the argument on why equal opportunity is better than demographic parity by~\citet{DBLP:conf/nips/HardtPNS16}: there is a difference between hiring equal numbers of males and females from ensuring that those numbers remain over time. 
In fact, this distinction summarizes why \citet{Wachter2020BiasPreserving} classifies fair ML definitions into bias preserving (like demographic parity) and bias transforming (like equal opportunity).

A second reason to consider the MM comparator is that, implicitly, we are already implementing methods that aim at ideal targets.
We are not proposing a drastic turn in fair ML research.
A good example of this point is the fairness-accuracy trade-off discussion \cite{DBLP:conf/nips/WickpT19, DBLP:conf/aies/SharmaZABMV20, DBLP:journals/natmi/RodolfaLG21, DBLP:journals/corr/abs-2011-03173, DBLP:conf/icml/DuttaWYC0V20}. 
Recent works on this topic agree that the trade-off exists as long as the data is biased. 
Hence, demonstrating its existence is trivial given the current state of our society; what is of interest is studying when or under which conditions the trade-off disappears.
These are hypothetical conditions that, though, serve as future targets to aim for.
Not surprisingly, in a setting in which $\mathbf{X}$ is non-informative of $A$ or, similarly, $A$ does not influence $\mathbf{X}$, there is no need to consider a trade-off between accuracy and fairness. 
In a fair world, the most accurate model is also the most fair. 
We believe that the MM comparator is meant to capture such world for the purpose of discrimination testing.

\subsection{Extending the Setup}
\label{sec:MM.GenerativeModels}

Let us consider indirect discrimination cases, meaning $A \rightarrow \mathbf{X} \rightarrow Y$.
Under the causal modeling problem of deriving a comparator, the key question to be answered is: \textit{what happens to $\mathbf{X}$ once we manipulate $A$ to derive the comparator?}
Given how Definitions \ref{def:TheStandardComparator} and \ref{def:MM-Comparator} differ, in the case of indirect discrimination the CP comparator ignores the downstream effect of $A$ on $\mathbf{X}$ while the MM comparator accounts for it.
However, while the implementation of the CP comparator is clear, how do we implement the MM comparator? 
% How do we define $\tilde{d}(\cdot, \cdot)$ in Equation~\eqref{eq:MMComparator}?
Answering this question helps extend the discrimination testing setup in Section~\ref{sec:ProblemFormulation.SetUp}. 
To start,
let $g(\cdot)$ represent a \textit{generative model} such that:
\begin{equation}
\label{eq:GenModel}
    \text{for} \;\; \tilde{\mathbf{X}} = g(\mathbf{X}, A) \;\;\; \text{we have} \;\;\; \tilde{\mathbf{X}} \perp\!\!\!\!\perp A
\end{equation}
We view $\tilde{X}$ \textit{as an adjusted representation of} $\mathbf{X}$ that is removed from the influence of the protected attribute $A$.
Conceptually, Equation~\eqref{eq:GenModel} describes the problem of fair representation learning \cite{Zemel2013LearningFairRepresentations} in which we want to learn or have access to $g(\cdot)$ to achieve $\tilde{\mathbf{X}}$.
Such a model, for example, can be a Variational Auto-Encoder or a Markov Chain \cite{DBLP:journals/corr/KingmaW13, DBLP:conf/icml/SalimansKW15, DBLP:conf/nips/BengioYAV13}.

For our purposes, we view $\tilde{\mathbf{X}}$ as a counterfactual (fair) representation of $\mathbf{X}$ that answers the question of what would have happened had the individual belonged to the non-protected group instead of the protected group. 
Hence, $g(\cdot)$ represents the counterfactual generation procedure of abduction, action, and prediction described in Section~\ref{sec:Causality}.
Given $g(\cdot)$, we should be able to generate the set of non-protected attributes of the complainant $i$'s comparator $i'$ and derive the MM comparator $i'$ such that:
\begin{equation}
\label{eq:TildeX}
    d ( \Tilde{\mathbf{x}}_i, \mathbf{x}_{i'} ) \leq \epsilon
\end{equation}
where possibly $\Tilde{\mathbf{x}}_i \neq \mathbf{x}_i$. Given the causal relation $A \rightarrow \mathbf{X}$, the comparator $i'$ is similar to the counterfactual profile of $i$ but dissimilar to its factual profile. 
Here, $d(\cdot, \cdot)$ represents again a distance function unaware of the protected attribute $A$ in Equation~\eqref{eq:TheStandardComparator}.
Through Equation~\eqref{eq:TildeX}, we summarize the link between the Definitions \ref{def:TheStandardComparator} and \ref{def:MM-Comparator} complainant $i$ and its comparator $i'$ as:
\begin{equation}
\label{eq:LinkBtwDefs}
    \tilde{d} \big( (\mathbf{x}_i, a_i), (\mathbf{x}_{i'}, a_{i'}) \big) = d \big( g(\mathbf{x}_i, a_i), \mathbf{x}_{i'} \big) 
\end{equation}
where the left-hand-side of the equation represents the general definition of the MM comparator while the right-hand-side an implementation of it under the generative model $g(\cdot)$.

The relationship in Equation~\eqref{eq:LinkBtwDefs} allows us to derive the MM comparator via $\tilde{d}(\cdot, \cdot)$ or via $g(\cdot)$ plus $d(\cdot, \cdot)$.
Our preference, based on our counterfactual reasoning approach, is with using $g(\cdot)$ plus $d(\cdot, \cdot)$, which allows for: (i) making clear the difference between the MM and the CP approaches; (ii) reusing the existing notions of distances and algorithms for inferring them \cite{DBLP:journals/ftml/Kulis13,DBLP:conf/forc/Ilvento20}. 
In that case, the setup in Section~\ref{sec:ProblemFormulation.SetUp} would require an additional element on top of deriving the comparator and comparing the outcomes: learning $g(\cdot)$, which is of course not limited to counterfactual generation.
This extended setup establishes a connection between discrimination testing and the field of fair representation learning \cite{DBLP:journals/corr/abs-2407-03834}, highlighting a promising direction for future research.

%
% EOS
%

\section{An Illustrative Experiment}
\label{sec:Experiments}

To showcase the CP (Definition~\ref{def:TheStandardComparator}) and MM (Definition~\ref{def:MM-Comparator}) comparators and their impact on discrimination testing, we consider the Law School Admissions example popularized by \citet{Kusner2017CF} and use on it the discrimination tools of situation testing (ST) \cite{Thanh_KnnSituationTesting2011} and counterfactual situation testing (CST) \cite{CST23, DBLP:journals/jair/AlvarezR25}.
ST and CST are both data mining tools that look for $k$-nearest neighborhoods for complainant-comparator pairs in a dataset.
We focus on these tools because CST extends ST by implementing the MM comparator over the CP comparator.
Both tools follow a similar pipeline, even sharing the same distance function for constructing the neighborhoods around the complainant and comparator, with the key difference that ST flips the protected attribute to create its (CP) comparator while CST uses the generated counterfactual of a complainant to create its (MM) comparator.
CST does so given the SCM $\mathcal{M}$ with DAG $\mathcal{G}$ shown in Figure~\ref{fig:LawSchool}. 
The code for this section is publicly available in a GitHub repository.\footnote{\url{https://github.com/cc-jalvarez/revisiting-the-comparator}.}

The dataset contains $n=21790$ applicants, 
containing the applicants' gender ($G$), race ($R$), undergraduate grade-point average ($UGPA$), and law school
admissions test score ($LSAT$). 
In terms of the protected attributes, 43.8\% of applicants are females and 16.1\% are non-whites.
Following \citet{CST23, DBLP:journals/jair/AlvarezR25}, we define a decision-maker $b(\cdot)$ that uses only $UGPA$ and $LSAT$ for admitting ($Y=1$) or not admitting ($Y=0$) an applicant. 
The decision-maker $b(\cdot)$ tests if an applicant's weighted sum of $UGPA$ and $LSAT$ is above the cutoff $\psi$, where $\psi$ is the median entry requirement of a top US law school.\footnote{The cutoff is the weighted sum of 60\% in $UGPA$ (3.93 over 4.00), and 40\% $LSAT$ (46.1 over 48), giving a total of 20.8. The maximum possible score given $b(\cdot)$ is 22. It is based on Yale University Law School: \url{https://www.ilrg.com/rankings/law/index/1/asc/Accept}.}
We are interested in testing, separately, whether female and non-white applicants are potentially discriminated against under this decision-making process.\footnote{Intersectionality, meaning what occurs with non-white-female applicants, is not considered under most non-discrimination laws \cite{Xenidis2020_TunningEULaw, Crenshaw1989_DemarginalizingTheIntersection, DBLP:journals/jair/AlvarezR25}. Testing-wise, it means that we only consider the claims for gender and race separately.}

CST proceeds as follows.
Given a tabular dataset of decisions, with each row denoting an individual, it detects whether each protected individual was or was not (potentially) discriminated against by the decision-maker.
Each protected individual represents a complainant $i$, and for each $i$ it generates its comparator $i'$ by running a counterfactual query on the protected attribute of interest (recall, Section~\ref{sec:Causality}) based on a SCM.
Using Gower's distance \cite{Gower1971}, CST mines for $k$-similar protected individuals to $i$ and $k$-similar non-protected individuals to $i'$.
The former represents the control group while the latter represents the test group.
Recall that $i'$ does not appear in the input dataset and answers the question \textit{what would have been of $i$ had it been a non-protected individual during the decision process?}
The control group is centered on $i$ while the test group is centered on $i'$.
CST then compares the difference in rejection proportions---meaning how many individuals out of $k$ in a group received a negative decision---between the control and test groups.
In its strictest form, CST considers as evidence for discrimination if the difference in proportions is strictly greater than zero, meaning the proportion of negative outcomes is greater in the control group compared to the test group.
Because it evaluates a difference in proportions, CST is able to equip each individual claim with confidence intervals. 
We do not use this aspect of CST here.
We include the relevant details of CST in Appendix~\ref{App:CST}.

ST follows similarly, with the key exception that it does not generate the comparator $i'$. 
Instead, it builds both control and test groups around $i$ using Gower's distance \cite{Gower1971}.
It means that $i'$, in practice, is the most similar non-protected individual to $i$ according to the distance function.
Both control and test groups are centered around $i$. 
For our purposes in this section, the rest of the procedure is similar to CST.
Importantly, ST excludes the complainant-comparator pair when looking at the difference in proportion of negative outcomes.
CST, instead, has the option to include or exclude the ``search centers''. 

\begin{figure}[t!]
\begin{minipage}{.35\linewidth}
\begin{figure}[H]
\centering
    \begin{tikzpicture}
        \node (A1)  at (-1.75, -0.75) [circle, draw]{R};
        \node (A2)  at (-1.75, 0.75) [circle, draw]{G};
        \node (X1) at (0, 1) [circle, draw]{UGPA};
        \node (X2) at (0,-1) [circle,draw]{LSAT};
        \node (Y)  at (1.5, 0) [circle, draw]{$Y$};
        \draw[->, thick] (A1) to (X1) {};
        \draw[->, thick] (A1) to (X2) {};
        \draw[->, thick] (A2) to (X1) {};
        \draw[->, thick] (A2) to (X2) {};
        \draw[->, thick] (X1) to (Y) {};
        \draw[->, thick] (X2) to (Y) {};
    \end{tikzpicture}
\Description{}
\end{figure}
\end{minipage}
\begin{minipage}{.55\linewidth}
\begin{align*}
\mathcal{M} \, & 
\begin{cases}
    R & \leftarrow U_{R}\\
    G & \leftarrow U_G \\
    UGPA & \leftarrow b_U + \beta_1 \cdot R + \lambda_1 \cdot G + U_1 \\
    LSAT & \leftarrow  b_L + \beta_2 \cdot R + \lambda_2 \cdot G + U_2 
    % LSAT & \leftarrow  \exp\{b_L + \beta_2 \cdot R + \lambda_2 \cdot G + U_2\} \\
\end{cases}
\end{align*}
\begin{align*}
Y & =
\begin{cases}
    1 \; \text{if} \; (0.6 \cdot UGPA + 0.4 \cdot LSAT) > \psi\\
    0  \; \text{otherwise} \\
\end{cases}
\end{align*}
\end{minipage}
\caption{The auxiliary causal knowledge with corresponding SCM $\mathcal{M}$ and DAG $\mathcal{G}$ for Law School Admissions (Level 3) \cite{Kusner2017CF}. 
Let $b_U$ and $b_L$ denote the intercepts and $\beta_1$, $\beta_2$, $\lambda_1$, $\lambda_2$ the regression weights.
% ; and $UGPA \sim \mathcal{N}$ and $LSAT \sim \text{Poisson}$ the probability distributions. 
These parameters are estimated from the data using ordinary linear regression. We obtain, respectively, the estimates $\hat{b_U}=3.21$, $\hat{\beta_1}=-0.22$, $\hat{\lambda_1}=0.13$ and $\hat{b_L}=37.8$, $\hat{\beta_2}=-4.64$, $\hat{\lambda_2}=-0.61$.}
\Description{}
\label{fig:LawSchool}
\end{figure}

To illustrate these two tools, consider a female applicant as the complainant $i$. 
ST finds $k$-females and $k$-males with the closest $UGPA$ and $LSAT$ to $i$ and compares the difference in the proportion of rejections between both groups. 
The complainant is considered discriminated against if such a difference is significantly different from zero.
The ideal comparator $i'$ in ST is a male with the same $UGPA$ and $LSAT$ as $i$.
CST does the same with the exception that it first generates $i$'s counterfactual via Figure~\ref{fig:LawSchool} and uses it to find $k$-males based on the closest $UGPA$ and $LSAT$. 
The female group is constructed in the same way as in ST as the counterfactual generation does not apply to them. 
This is because the counterfactual query of interest (\textit{what would have happened to the female applicant had she been male?}) is concerned with the group of male applicants.
There is no ideal comparator $i'$ to be found under CST but, instead, a counterfactual representation of $i$ in the form of $i'$ with $\Tilde{\mathbf{X}}$ \eqref{eq:TildeX}.

\begin{table}[t]
  \caption{Number (and \% over the total cases) of discrimination cases based on race.}
  \label{table:k-results_RACE}
  \centering
  \begin{tabular}{lccccc}
    \toprule
    Method & $k=25$ & $k=50$ & $k=100$ & $k=200$ & $k=500$ \\
    \midrule
    ST & 40 (1.14\%) & 61 (1.74\%) & 64 (1.83\%) & 75 (2.14\%) & 108 (3.08\%)\\
    CST & 296 (8.44\%) & 337 (9.61\%) & 400 (11.41\%) & 476 (13.58\%) & 605 (17.26\%) \\
    \bottomrule
  \end{tabular}
\end{table}
\begin{table}[t]
\caption{Number (and \% over the total cases) of discrimination cases based on gender.}
  \label{table:k-results_GENDER}
  \centering
  \begin{tabular}{lccccc}
    \toprule
    Method & $k=25$ & $k=50$ & $k=100$ & $k=200$ & $k=500$ \\
    \midrule
    ST & 117 (1.23\%) & 229 (2.4\%) & 258 (2.71\%) & 368 (3.86\%) & 492 (5.16\%) \\
    CST & 106 (1.11\%) & 253 (2.65\%) & 296 (3.1\%) & 449 (4.71\%) & 605 (6.34 \%) \\
    \bottomrule
  \end{tabular}
\end{table}

Because of how each method constructs the male neighborhoods, ST and CST implement, respectively, the CP and MM comparators.
Tables~\ref{table:k-results_RACE} and \ref{table:k-results_GENDER} show the number of \textit{prima facie} discrimination cases identified under ST and CST for, respectively, the protected attributes race (with the goal of answering \textit{what would have been the outcome had the applicant been white?}) and gender (and \textit{what would have been the outcome had the applicant been male?}) for different $k$ neighborhood sizes.
CST generates a counterfactual distribution for each query.
The CST results are without the complainant-comparator pair. This is because ST does not include them. We run CST without these ``search centers'' to have comparable results between both methods.

\begin{figure}[t]
    \centering
    \begin{subfigure}{.45\linewidth}
    \centering
    \includegraphics[scale=0.495]{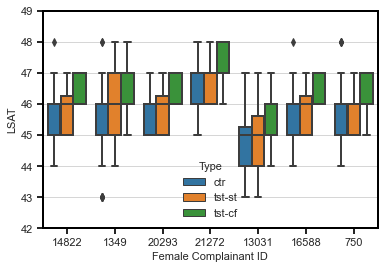}
    \end{subfigure}
    \hfill
    \begin{subfigure}{.45\linewidth}
    \centering
    \includegraphics[scale=0.495]{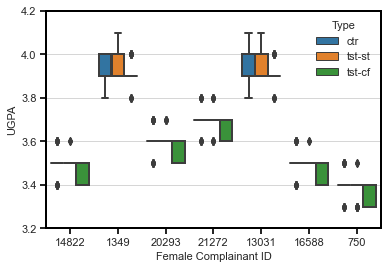}
    \end{subfigure}
\caption{We show the distributions in the form of box-plots for the $LSAT$ and $UGPA$ for seven randomly chosen complainants based on the protected attribute gender. These complainants are found to be discriminated by both ST and CST for $k=100$. The so-called control (ctr) are the neighborhoods of similar female profiles to the complainant, which is the same for both tools. Similarly, the so-called test groups (tst-st for ST; tst-cf for CST) are the neighborhoods of similar male profiles to the comparator: i.e., the CP comparator for ST, and the MM comparator for CST. In both figures we observe a difference between the two kinds of test groups. As we would expect, it means that the test and control groups are more similar under ST than under CST. Both figures show the impact of the CP and MM comparators.}
\Description{}
\label{fig:gender_analysis}
\end{figure}

\begin{figure}[t]
    \centering
    \begin{subfigure}{.45\linewidth}
    \centering
    \includegraphics[scale=0.495]{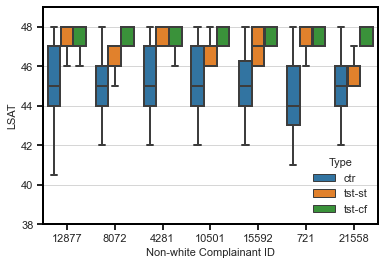}
    \end{subfigure}
    \hfill
    \begin{subfigure}{.45\linewidth}
    \centering
    \includegraphics[scale=0.495]{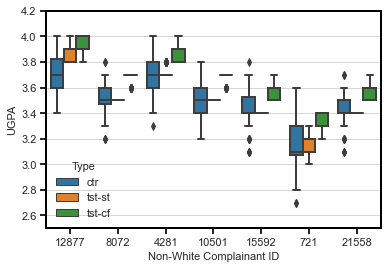}
    \end{subfigure}
\caption{The same analysis follows as in Figure~\ref{fig:gender_analysis} but based on the protected attribute race, including similar insights and conclusions on the influence of the CP and MM comparators. In both figures, we observe a difference between the two kinds of test groups, highlighting again the impact of the CP and MM comparators.}
\Description{}
\label{fig:race_analysis}
\end{figure}

The tables illustrate the impact of choosing between the CP and MM comparators, with CST detecting a higher number of \textit{prima facie} discrimination cases for the same dataset.
\textit{This difference is driven by ST, under the CP comparator, and CST, under the MM comparator, aiming, respectively, for an idealized and a fairness given the difference comparison.}
Intuitively, under a stricter or more conservative notion of similarity, ST constructs similar neighborhoods based on the protected attribute.
For instance, we observe this difference in Figure~$\ref{fig:gender_analysis}$ that shows seven randomly chosen individuals considered to have been discriminated against by both ST and CST on the basis of their gender. 
We compare the $LSAT$ and $UGPA$ distributions in the ``test groups'' (i.e., the neighborhoods constructed around the comparator) of ST and CST to the $LSAT$ and $UGPA$'s distribution in the ``control group'' (i.e., the neighborhood constructed around the complainant), which is the same for both ST and CST.
The two groups under ST for $LSAT$ and $UGPA$ are similarly distributed while this is not the case under CST. 
Figure~\ref{fig:gender_analysis} illustrates $\Tilde{\mathbf{X}}$ \eqref{eq:TildeX} and its role in discrimination testing.
In Figure~$\ref{fig:race_analysis}$, we carry out the same analysis and find similar results for race.

The results are starker for race than for gender. 
This point is observed by the difference between ST and CST for each protected attribute across $k$ size. For instance, at $k=50$, CST almost triples ST in number of detected cases for race but shows just a marginal increase for gender.
What is clear from these results is that the effect of race and gender on $LSAT$ and $UGPA$, within at least the context of law school admissions, is not symmetrical---as also confirmed by the parameter estimates reported in the caption of Figure \ref{fig:LawSchool}.
As argued in other works \cite{Fryer2010AnalysisOfActingWhite, Austen2003EconomicsOfActingWhite}, each protected attribute conditions its members differently as each protected attribute came to be, at least under a social construct perspective \cite{Mallon2007SocialConstruction}, with different purposes.
Based on these results and our conceptions of the CP and MM comparators, the similar results between ST and CST for gender would indicate a less pronounced impact from gender on $LSAT$ and $UGPA$. 
In other words, not much changes when generating the counterfactual profiles. The opposite occurs with race.
These results seem in line with work on how race can be more detrimental than gender in US college admissions \cite{Baker2018RaceAndStratification, Conger2013GenderGaps}.
Further, these results point to CST's capacity to capture less pronounced forms of discrimination.
Be that as it may, both tables do show evidence of potential discrimination by the decision-maker $b(\cdot)$, with CST detecting more cases.
As discussed in Section~\ref{sec:ProblemFormulation.Discrimination}, detecting discrimination is one aspect of establishing discrimination. Indeed, CST paints a worse setting than ST for the admissions process in question, but, regardless, the results of ST or CST would then have to be argued among stakeholders (such as the university itself). It varies on a case-by-case basis and the university can always provide, in principle, an objective justification.

The complainant used by CST for drawing the so-called test groups is generated under counterfactual generation using the abduction, action, and prediction steps \cite{Pearl2016_CausalInference}.
These profiles determining which male (or white) outcomes are to be compared to the female (or non-white) outcomes are purely hypothetical.
\textit{Each of these generated profiles represents what would have been of the female (or non-white) individual had it been male (or white) in the admissions process to law school.} 
In generating them and, in turn, adjusting for the impact of gender (or race) on $UGPA$ and $LSAT$, CST creates individual targets for each protected individual that represent where they would have been in a fairer (read, free from the impact of the protected attribute) world.
CST tests for equality under these targets.

These results are specific to ST and CST and the potential limitations inherent to both methods, though they still highlight what it would mean to test for discrimination under the MM comparator instead of the CP comparator.
Notably, the notion of generating the comparator is non-trivial. A method like ST is more conservative in its treatment of similarity between protected and non-protected individuals but, at the same time, it is also more intuitive.
The CP comparator is the standard because it offers a simple view on similarity.
A method like CST, instead, enters into a more subjective realm of what similarity should be. 
Although it relies on a modeling counterfactual approach in which the generated counterfactual distribution is unique to the given SCM and the specific intervention, said distribution that supports the instances used as comparators is one of other plausible counterfactual representations of how the world should be \cite{Woodward2005MakingThigsHappen}.
Some representations are more plausible than others, but still it is not as though CST builds on a shared ground truth of what the world should be.
The MM comparator, in that regard, poses a more challenging and, even, uncomfortable challenge of picturing in practice the sort of equality we want to aim for \cite{Wachter2020BiasPreserving}.

Further, in the case of CST, it is worth noting that the derivation of the SCM in Figure~\ref{fig:LawSchool} is a difficult task. 
For our purpose, which is to showcase the CP versus MM comparison, this limitation is taken for granted.
However, in practice, it requires either agreeing on the SCM for a given context, or discovering it based on data and knowledge using causal discovery techniques \cite{DBLP:journals/csur/VowelsCB23,DBLP:journals/widm/NogueiraPRPG22}.
\citet{DBLP:journals/jair/AlvarezR25} provide a discussion on this limitation, which, in short, can be summarized as a pragmatic view on causal modeling \cite{DBLP:conf/icml/Loftus24}: the SCM help to explicitly state our assumptions about the world. That is what makes them useful in settings in which disagreement among stakeholders is likely, such as discrimination testing.
Additionally, CST depends on a SCM that describes the dataset and the complexity of the dataset used will, thus, condition the SCM. 
Similarly, CST depends on Gower's distance function.
Here, we did not explore these limitations.
These limitations are independent of the MM comparator as a concept, though they should be taken into account in case we wish to implement it using SCM.

%

%
% EOS
%

\section{Overview of Discrimination Testing Tools}
\label{sec:Survey}

In this section, we survey a representative set of tools for discrimination testing focusing on the comparator.
Different from Section~\ref{sec:RelevantRW}, the goal is not to position this work but rather to provide an overview of the methods, which span multiple fields, while using our novel formalization of the \textit{ceteris paribus} and \textit{mutatis mutandis} comparators.
We recommend \citet{Romei2014MultiSurveyDiscrimination} for an extensive and multidisciplinary survey.
All these tools amount to gathering \textit{prima facie} evidence against a decision-maker, meaning they derive the comparator $i'$ for the complainant $i$ and compare their outcomes.
We view these tools as instances of the counterfactual model of discrimination.

\subsection{Standard Tools}

Here, by standard we mean non-algorithmic tools. 
These tools mainly include natural experiments (e.g., \citet{Godin2000Orchestra}), field experiments (e.g., \citet{Bertrand2017_FieldExperimentDiscrimination}), and audits (e.g., \citet{Fix&Struyk1993_ClearConvincingEvidence}). 
These tools share a focus on either gathering or evaluating data that is, by design, able to capture the effect of the protected attribute on the decision outcome.
Such data is characterized by a before-and-after or a with-and-without mechanism around the protected attribute that ensures the identifiability of the protected attribute's effect.
These tools are somewhat outdated for how decisions are taken in modern decision flows, in which ML may enable and even automate key steps.

Let us start with audits, which are a more holistic kind of tool, very human-dependent, and less focused on modeling. 
Audits consist of examining the decision process in question through human auditors. 
In practice, it means sending a group of experts to observe during a period of time the decision-maker and study its practices to corroborate the claims of the complainant \cite{Fix&Struyk1993_ClearConvincingEvidence}.
% Auditors, e.g., tend to interview key stakeholders.
In the case of the modern decision flow, such a practice only makes sense if the model is used by a human decision-maker or if the focus is to audit the procedure by which the model was trained \cite{doi:10.1098/rsos.230859}, which is increasingly being viewed as a priority over interpretability by regulators \cite{DBLP:conf/fat/PaniguttiHHLYJS23}.
The focus is on understanding the overall process and identifying procedural concerns. The comparator is not an explicit objective.

Natural and field experiments, instead, have more of a focus on discrimination as a modeling problem (i.e., identifying the causal effect of the protected attribute) and, thus, prioritize the designing and gathering of data suitable for this purpose.
Notably, these methods are a product of their time. 
In particular, these methods are a product of the so-called ``empirical revolution'' experienced by the social sciences in which a mixture of data availability, faster and cheaper computers, and an intuitive causal framework---the Potential Outcomes Framework---made it possible to test causal claims via experiments \cite{Angrist2008MostlyHarmless}.

The idea behind natural experiments is simple.
A policy introduction or external shock occurs, creating a before-and-after scenario that, in principle, allows us to test for the effect of the protected attribute. 
We, thus, find ourselves naturally in an experimental setting.
\citet{Godin2000Orchestra}'s \textit{Orchestrating Impartiality: The Impact of ``Blind'' Auditions on Female Musicians} is a clear example.
US orchestras before the 1970s used to audition musicians without screens, meaning musicians would perform in full sight to the evaluating committee. 
The US orchestra system back then was male-dominated and openly sexist, which motivated the suspicion that there was discrimination against female musicians. 
Orchestras then started to conduct blind auditions to ensure an impartial process. 
Some orchestras even introduced carpets so that female musicians would not be identified by the sound of their heels when walking up the stage.
Female representation in US orchestras increased after the introduction of the screen.
\citet{Godin2000Orchestra}, though, using data from the auditions, set out to study whether this amounted to evidence for gender discrimination within the US orchestra system. 
There was indeed a before-and-after mechanism through the introduction of the screen that coincided with an increase in female musicians in orchestras, but to claim a causal effect it was necessary to control for other factors.
For instance, the introduction of the screen also coincided with an overall increase in female participation in the US workforce.
\citet{Godin2000Orchestra} used a regression model to control for this and other factors, finding inconclusive evidence for discrimination against female musicians. 
\citet{Godin2000Orchestra} is a seminal paper, among other things, because it shows how meticulous the model choice has to be to claim discrimination evidence.
The paper illustrates the importance of deriving a comparator good enough to isolate the effect of the protected attribute.

Field experiments, instead, focus on designing an experiment that creates the data for testing the discrimination claim.
The most famous example is \citet{Bertrand2004_EmilyAndGreg}'s \textit{Are Emily and Greg More Employable Than Lakisha and Jamal? A Field Experiment on Labor Market Discrimination}. 
Also see \citet{Bertrand2017_FieldExperimentDiscrimination} for a survey on field experiments for discrimination.
\citet{Bertrand2004_EmilyAndGreg} proposed the following experiment: let us create pairs of fictitious CVs tailored to job positions that are the same except for the candidate's name, where the name is intended to reflect both gender and race within the US context. 
They would create, for example, the same CV for Mike and Jamal (respectively, a common white and black name according to census data at the time); send both CVs to the same job openings via mail (this was done in the early 2000s); and compare which CV received a callback. 
The CV with the black-sounding name was the complainant and the CV with the white-sounding name its comparator.
All the CVs were drafted by the researchers. In the end, they found evidence for discrimination. This sort of field experiment is known as a \textit{correspondence study}. 
See \citet{Rooth2021} for a survey on correspondence studies.
Another kind of field experiment is \textit{situation testing} \cite{Rorive2009_ProvingDiscrimination} in which, rather than sending CVs or other forms of correspondence, we send a group of actors that share key physical traits and backstories as the complainant and its comparator.

Standard tools are at the center of the criticism against the counterfactual model of discrimination. 
This fact is not surprising, as these tools have been around longer, even inspiring the recent algorithmic tools in the next sections.
They represent the starting point for more novel discrimination testing methods, such as situation testing \cite{Thanh_KnnSituationTesting2011} and counterfactual situation testing \cite{DBLP:journals/jair/AlvarezR25}, both proposing an algorithmic version of the human-based situation testing \cite{Rorive2009_ProvingDiscrimination}.
\citet{Heckman1998_DetectingDiscrimination}, for example, is critical of the effectiveness of these methods in controlling for similarity and other factors and argues that we have been too confident in claiming to be able to detect discrimination.
More recently, as we discussed in earlier sections, \citet{Kohler2018CausalEddie} attacks these methods (she is particularly critical of \citet{Bertrand2004_EmilyAndGreg}) due to their experimental designs and causal treatment of the protected attribute. 

These standard tools rely on the CP comparator. 
As discussed previously, under direct discrimination, these tools are still useful and still have a lot to offer to algorithmic tools in terms of their focus on inferential statistics.
If, however, we are interested in proving indirect discrimination, then these tools and their experiments need to make a stronger argument for using the CP comparator. 
In particular, this argument should be centered on the kind of signal, as in relevant information, being used by the decision-maker and whether that signal has any link to the protected attribute in question.
Consider the case of the orchestra auditions \cite{Godin2000Orchestra}. 
There, the decision-maker cares only for the playing quality of the musician, which should not be related to gender.\footnote{Though there is a reasonable case to be made that instruments tend to favor male musicians \cite{CriadoPerez2019InvisibleWomen}.}
Hence, the signal can be recovered once the screen is imposed as the sounds can go through the screen but all other factors, like the musician's appearance, are blocked.
In such a setting, the CP comparator might already be well suited.
Consider now the case of the CVs \cite{Bertrand2004_EmilyAndGreg}. 
There, even if we impose a ``screen'' by removing the names, there is still the possibility that the decision-maker uses signals for candidate potential that are linked to race or gender, such as extracurricular activities or the name of the high school. 
These are valid signals, or at least signals that could be well argued by the decision-maker as necessary for measuring candidate potential.
In such a setting, the MM comparator might be better suited.

\subsection{Discrimination Discovery}

The discrimination discovery tools \cite{DBLP:journals/kais/KamiranZC13, Ruggieri2010_DMforDD, DBLP:conf/kdd/PedreschiRT08} represent the first attempt at what we now call algorithmic fairness.
These works took the standard methods of the previous section and combined them with data mining methods, shifting the focus from generating or finding the experimental data to exploring data already produced by a decision process. 
This shift in focus was in part field-driven: within data mining and overall knowledge discovery, the focus of the problem formulation is on developing new and efficient algorithms for extracting information from large datasets. 
It is a different focus from the prevalent inferential statistics approach in which the creation of the data is as important as using the most efficient model.
But this shift also came with a change in how we test for discrimination, especially with the adoption of modern decision flows,
% not convinced by this part:
as the attention is now on a consistent, algorithmic decision-maker.
Hence, to test whether the decision-maker discriminated, we could now look at the data it produced. 

The setup in Section~\ref{sec:ProblemFormulation.SetUp} started to become standard under these tools.
It was a natural transition from the standard tools as these works borrowed considerably both from non-discrimination law and from the standard methods. 
For example, works by \citet{DBLP:conf/kdd/PedreschiRT08} and \citet{Ruggieri2010_DMforDD} provided the first formalization of equal opportunity (and, implicitly, of equalized odds) considerably before \citet{DBLP:conf/nips/HardtPNS16}. 
\citet{Kamiran2009_ClassifyingWihtoutDiscriminating} did the same for demographic parity.
\citet{Thanh_KnnSituationTesting2011} is an illustrative example of the discrimination discovery wave, proposing a $k$-NN implementation of situation testing \cite{Rorive2009_ProvingDiscrimination}. 

These tools largely focus on defining the best distance to reach the comparator for a given complainant.
Some of the tools are closer to the MM comparator, but most of them implement the CP comparator.
\citet{Thanh_KnnSituationTesting2011} clearly use the CP comparator as they minimize the distance between the complainant and potential comparators based only on the set of non-protected attributes.
\citet{Zhang_CausalSituationTesting_2016}, instead, offer an extension to \cite{Thanh_KnnSituationTesting2011} via causal weights, focusing on defining a distance function that prioritizes the important non-protected attributes when finding the comparator for a complainant.
Similarly, \citet{DBLP:journals/jiis/QureshiKKRP20} use a propensity score weight for the distance function with the protected attribute as the treatment. 
None of these tools, however, explicitly address the problem with using the CP comparator or the need to derive the MM comparator.
For example, there is no discussion on how the protected attribute might influence all other attributes.

\subsection{Algorithmic Fairness}

Discrimination discovery took a broader, more multidisciplinary focus over the past decade and became algorithmic fairness.
The objectives are the same, but algorithmic fairness is broader in scope, usually aiming to reformulate standard ML problems under a fairness objective \cite{Barocas2019FairML}.
In turn, many of these tools, even if they are not intended to prove discrimination, are useful for tracking potentially discriminatory decision-making.
Unsurprisingly, the terms unfair and discriminatory are used interchangeably with algorithmic fairness, sometimes incorrectly \cite{DBLP:conf/fat/WeertsXTOP23}. 
As discussed in Section~\ref{sec:ProblemFormulation.Discrimination}, discrimination claims must align with non-discrimination law in terms of the protected attribute and testing tools represent one aspect of establishing discrimination.

Given the counterfactual model of discrimination and our focus on the comparator, the main algorithmic fairness works that are central to discrimination testing are counterfactual fairness \cite{Kusner2017CF} and individual fairness \cite{DworkHPRZ12}. 
The former establishes the same treatment of the factual and its counterfactual instance under the same decision; the latter establishes the same treatment of similar instances under the same decision.
Both fairness definitions allude explicitly to the complainant-comparator pair.
Under the counterfactual model of discrimination, these two concepts intersect as we use the similarity between individuals to approximate the comparison between an individual and its counterfactual-self. 
Works like the FlipTest \cite{BlackYF20_FlipTest} or CST \cite{CST23} build upon these two definitions. Overall, this holds for other similar works including \cite{DBLP:journals/tit/DuttaVMDG21, DBLP:conf/nips/AnthisV23, DBLP:conf/sigmod/SalimiRHS19}.
Now algorithmic fairness is a vast and ongoing field, with research on fairness definitions and their links to non-discrimination law being one active branch \cite{Wachter2020BiasPreserving, Wachter2021FairnessCannotBeAutomated}.
Moving forward, it is important that the CP versus MM comparator discussion is present when looking at (algorithmic) discrimination.
The choice of method or definition matters beyond providing a complete view when running experiments. 
Such a choice, as discussed, has wider consequences on what we consider discriminatory that are normative concerns rather than modeling concerns.
Each method, as also argued by \citet{Wachter2020BiasPreserving}, operationalizes a specific ML ``intervention'' to the systematic unfairness behind the discrimination claim.
Of this modeling choice we must be aware.

Moving forward, especially given our formulation of $\Tilde{\mathbf{X}}$ in Section~\ref{sec:MM.GenerativeModels},
we view algorithmic fairness tools based on generative models as central to the MM comparator. 
In particular, methods concerned with learning fair representations. This research direction started with \citet{Zemel2013LearningFairRepresentations} and has led to multiple proposed methods over the past decade (e.g., \citet{DBLP:conf/aaai/CerratoKE023, DBLP:journals/corr/abs-2201-06336, DBLP:journals/jmlr/GaninUAGLLML16, DBLP:journals/corr/LouizosSLWZ15, DBLP:conf/nips/MoyerGBGS18, DBLP:conf/nips/XieDDHN17, DBLP:conf/icml/MadrasCPZ18}).
The goal with these methods is to learn a fair representation of the individual, meaning learning a representation of seemingly neutral attributes that removes information from the protected attribute while keeping valuable information of the attributes in question useful for a downstream task.
We recommend \citet{DBLP:journals/corr/abs-2407-03834} for a survey on fair representations.
Since those complex models tackle the problem (\ref{eq:GenModel}), they could be, in principle, repurposed for discrimination testing with the goal of deriving an MM comparator. Future research is open along this interesting line. Our approach relies on SCM, for which such a derivation boils down to counterfactual generation (cfr. Section~\ref{sec:Causality}).

Algorithmic fairness tools that are more focused on proving discrimination, meaning along the same lines as discrimination discovery, still face the key challenge of linking unfairness to discrimination \cite{DBLP:conf/fat/WeertsXTOP23}. 
These tools are useful, but it remains to be seen how they fit into the pipeline for establishing discrimination. This is not a technical challenge in itself, but one on whether the technology is embraced or not by the relevant stakeholders.
Another key challenge, which links directly to the MM comparator, is how these tools address indirect discrimination as most algorithmic decision-makers are not allowed to use the protected attribute as input.
As illustrated in Section~\ref{sec:Experiments}, the choice of comparator affects the number of discrimination cases identified.
These tools, along with the discrimination discovery ones, are capable of generating more complex comparators than the standard tools. 
The question then is whether or not they should aim at implementing the MM comparator over the CP comparator. 
Our view is, at a minimum, they should allow for the possibility of choosing between the two.

%
% EOS
%

\section{Conclusion}
\label{sec:Conclusion}

This work discussed the problem of discrimination testing by focusing on the comparator.
First, we presented the derivation of the comparator as a causal modeling problem. 
It is not the most popular view on discrimination testing \cite{Kohler2018CausalEddie}, but an important one as it opens impactful opportunities for ML models.
Second, we introduced the \textit{ceteris paribus} (CP) and \textit{mutatis mutandis} (MM) comparator distinction.
The former refers to the standard comparator and the latter represents a new kind of comparator that departs from idealized comparisons.
This distinction is novel and will help not only to classify existing methods but also to better inform new ones and how their choice of similarity between individuals unavoidably impacts normative concerns related to discrimination.
Additionally, the illustrative experiment based on situation testing \cite{Thanh_KnnSituationTesting2011} and counterfactual situation testing \cite{CST23, DBLP:journals/jair/AlvarezR25} showcased the impact of this distinction on discrimination testing, motivating the need to always have the option of choosing the kind of comparator when implementing a discrimination testing tool.

Here, we focused on causal modeling, particularly structural causal models (SCM) for generating the MM comparator.
We stress that the MM comparator is not tied to this modeling choice and other ML approaches are possible and encouraged. 
Fair representation learning \cite{DBLP:journals/corr/abs-2407-03834} is the most appealing direction in terms of generative ML for the MM comparator. 
That said, given our causal modeling choice for generating the MM comparator, two limitations are clear and worth discussing.
First, we use the same procedure as in \citet{Kusner2017CF} to generate the suitable imaginary comparators, meaning using \citet{Pearl2016_CausalInference}'s abduction, action, and prediction steps. 
It remains a common procedure that can rely on either a frequentist or Bayesian approach.
However, it is also a procedure that requires sufficient knowledge about the SCM generating the data, which is complex and potentially non-scalable in practice.
Other causal ML approaches, such as causal normalizing flows \cite{DBLP:conf/nips/JavaloySV23}, offer an alternative to generating the counterfactual distribution without having to be explicit about model specifications in the SCM.
We do not explore these methods as our focus is on the CP versus MM comparison.\footnote{We also take for granted, as most SCM-based works, that there is only one SCM to consider in our testing context. Competing SCMs, for instance, can introduce causal perception and influence the call on whether there is evidence of discrimination. See \citet{Alvarez2025CausalPerception} for details.}
We leave this as future work.
Second, our extension of the standard discrimination testing setup (Section~\ref{sec:MM.GenerativeModels}) centers on generating an adjusted representation of the non-protected attributes.
In that sense, it aligns well with the problem of fair representation learning. 
However, an alternative approach could be to define an adjusted similarity function with the same objective.
We do not consider this approach as, given our use of SCM and overall emphasis on generative ML, the generation of $\Tilde{\mathbf{X}}$ fits better.
We too leave this as future work. Exploring different kinds of data-driven distance functions \cite{DBLP:journals/pr/BlancoMalloMRB23} as well as searching for similar individuals based on their latent space representation are potential directions to explore.

%Overall, 
Defining a distance function can be as challenging as agreeing on a SCM for generating the MM comparator in the form of $\Tilde{\mathbf{X}}$.
Consider, for instance, the discussions on implementing individual fairness by \citet{DworkHPRZ12}, which always come back to the circularity argument in Section~\ref{sec:ProblemFormulation} inspired by \citet{Westen1982EmptyEquality}.
Defining a similarity measure between individuals is inherently contentious. 
At least with the SCM our answer to the \textit{what would have been if} question is answered by the counterfactual distribution and, in turn, settles the ``choice of similarity metric'' debate as a generative question. Redirecting the focus to a distance function, we believe, would still require defining an imagined objective to learn the distance function.
In fact, said distance could be based on evaluating the factual and counterfactual distributions. This is a promising line of research.

Further, given now the proposed CP and MM comparator distinction, a clear direction for future work is to move beyond an illustrative experiment and survey suitable methods for the MM comparator in a robust experimental setting.
The goal here was to motivate and formalize such distinction, which is by all means not trivial and often not shared among all of those in the discrimination testing field.
Future work should move beyond SCM and consider the methods already mentioned as well as other fair ML benchmark datasets to explore the potential limitations of the MM comparator as a modeling problem.

Finally, this work comes from a ML perspective on discrimination testing. 
We have done our best in incorporating other fields and their views on the need for the MM comparator.
In doing so, we focused on one aspect---that of providing \textit{prima facie} evidence---of the full pipeline for establishing discrimination.
We hope future legal work complements what we have presented here as the choice of a MM over a CP comparator would already expand the current pipeline, requiring additional legal commentary.
We are certain that the MM comparator and its departure from the CP comparator, both as a conceptual and modeling problem, will motivate further fair ML work on discrimination testing.
As shown in Example~\ref{ex:RunningExample}, it is not straightforward to determine who Clara should be compared to when evaluating her discrimination claim. The proposed MM comparator helps facilitate this complex and important discussion.

%
% EOS
%

%%
\begin{acks} 
    We would like to thank Vijay Keswani, Otto Sahlgren, Mattia Cerrato, Alesia Vallenas Coronel, Marcello Di Bello, Nicolò Cangiotti, Francesco Nappo, and Michele Loi for their feedback at different stages of this work. This work has received funding from the European Union’s Horizon 2020 research and innovation program for the MSCA project NoBIAS (g.a. No. 860630) and for the project FINDHR (g.a. No. 101070212). It reflects only the authors' views and the European Research Executive Agency is not responsible for any use that may be made of the information it contains.
\end{acks}

\appendix
\section{On \textit{The Trouble with Disparity}}
\label{App:Disparity}

Another critical work toward discrimination testing worth mentioning is Michaels and Reed's article, \textit{The Trouble with Disparity} \cite{TheTroubleWithDisparaty}. 
We believe that it offers a similar critique to \citet{Kohler2018CausalEddie}, but from a non-legal angle.
Michaels and Reed, in particular, focus on the tension between race and class and use the coverage given to black communities during the COVID-19 pandemic to illustrate their points.
During the pandemic in the USA, blacks were dying from the virus at a higher rate than any other social group, which prompted ``scientific'' questions on whether there was a genetic predisposition that made this group more vulnerable to the virus relative to other groups.
No such evidence was found.
Once we controlled for other factors to isolate the so-called treatment effect of race, it became clear that ``race'' had nothing to do with the likelihood of dying from the virus.
Drawing parallels to non-discrimination law and its focus on parity based on representation, \citet{TheTroubleWithDisparaty} point to the obvious or, in their words, deeper cause that nobody wanted to address regarding the disparate death rates among social groups from COVID-19: inequality.
Blacks were not more likely to die from COVID-19 because of the color of their skin but because of their socioeconomic background.
Being poor is what made an individual in the USA more vulnerable to COVID-19, and the majority of poor individuals in the USA are black.

We mention this article because it highlights the complexity of discrimination testing.
The answer to ``why are most low-income individuals in the USA today black?'' is the same answer to ``why is race considered a protected attribute?'': because there is a history of systematic policies against individuals perceived as black (or, in general, non-white). 
Now, because of that same answer, the question we aim to answer when testing for the discriminatory effects of race should not be limited to race alone. 
The same way we have established that race and other protected attributes exist and have acknowledged their effects in our society, we must also recognize how their effects have materialized in limiting the opportunities of multiple generations.
Back to the COVID-19 example, it might have been race what initially caused the exclusionary policies toward non-whites that, in turn, created the current social context, but, precisely because of this historical process, race alone cannot claim full responsibility for the present disparities.
Larger forces, though, still associated with race, such as inequality and social mobility, are stronger causes of the present disparities. 
See, for example, \citet{Chetty2020_Race}.
The problem is that, ironically, as Michaels and Reed argue, it has become easier to speak of racial discrimination than income discrimination.

\section{Supplementary Material for Counterfactual Situation Testing}
\label{App:CST}

In this section, we present additional technical material for CST \cite{CST23, DBLP:journals/jair/AlvarezR25}. CST relies on the k-NN algorithm, building the neighborhoods of the control and test groups using a distance function $d(\cdot,\cdot)$ between two individual profiles or tuples. Such function only looks at non-protected attributes.
Let us define the between tuple distance $d(x_1, x_2)$ as:
\begin{equation}
\label{eq:Distance}
    d(x_1, x_2) = \frac{\sum_{i=1}^{|X|} d_i(x_{1, i}, x_{2, i})}{|X|}
\end{equation}
such that $d(x_1, x_2)$ averages the sum of the per-attribute distances $d_i(x_{1,i}, x_{2, i})$ across all attributes in $X$.
CST handles non-normalized attributes but, as default, we normalize them to ensure comparable per-attribute distances.

The $d_i$ used depends on the type of the \textit{i-th} attribute.
It equals the overlap measurement ($ol$) if the attribute $X_i$ is categorical; otherwise, it equals the normalized Manhattan distance ($md$) if the attribute $X_i$ is continuous, ordinal, or interval.
Hence, $d(\cdot,\cdot)$ amounts to Gower's distance \cite{Gower1971}.
We recall $md$ and $ol$ distances below:
\begin{equation}
    md(x_{1,i}, x_{2, i}) = \frac{| x_{1,i} - x_{2, i} |}{(\max(X_i) - \min(X_i))}
\end{equation}
\begin{equation}
    ol(x_{1,i}, x_{2, i}) = 
    \begin{cases}
    1 & \text{if } x_{1, i} \neq x_{2, i} \\
    0 & \text{otherwise}.
\end{cases}
\end{equation}
The main objective of CST is to estimate the difference in proportion of negative decision outcomes for each complainant's control and test groups:
\begin{equation}
\label{eq:delta}
    \Delta p = p_c - p_t
\end{equation}
where $p_c$ and $p_t$ represent the count of individuals with a negative decision outcome, respectively, in the control group and test group.
Recall that the control group is centered on the complainant $i$ and the test group on its (generated counterfactual) comparator $i'$.

For our purposes, there is potential individual discrimination toward a complainant if $\Delta p > \tau$, where $\tau \in [-1, 1]$ represents some accepted deviation. Often, $\tau=0$. This would amount to a general \textit{prima facie} discrimination claim. CST can also equip such claim with confidence intervals, addressing concerns on whether it is statistically significant or not. For more details, we invite the reader to see \citet{DBLP:journals/jair/AlvarezR25}.

%
% EOF
%

%%
\bibliographystyle{ACM-Reference-Format}
\bibliography{references}

\end{document}